\documentclass{article}

    \PassOptionsToPackage{comma}{natbib}

\usepackage[main, final]{neurips_2026}

\usepackage[utf8]{inputenc} 
\usepackage[T1]{fontenc}    
\usepackage[backref = page, colorlinks = True, linkcolor = violet!70!white, citecolor = teal]{hyperref}       
\usepackage{url}            
\usepackage{booktabs}       
\usepackage{amsfonts}       
\usepackage{nicefrac}       
\usepackage{microtype}      
\usepackage[dvipsnames,svgnames,x11names]{xcolor}
\usepackage{multirow}
\usepackage{multicol}
\usepackage{bm}
\usepackage{subfigure}
\usepackage{enumitem}

\usepackage{amsmath,amssymb,amsthm}
\usepackage{mathtools}
\usepackage{algorithm}
\usepackage{wrapfig}
\usepackage{cleveref}
\usepackage{algpseudocode}  
\usepackage{makecell}
\usepackage{pifont}
\usepackage{makecell}
\usepackage{xspace}

\newcommand{\Dcal}{\mathcal{D}}
\newcommand{\Xcal}{\mathcal{X}}
\newcommand{\Ycal}{\mathcal{Y}}

\newcommand{\vpmse}[2]{$#1 \textcolor{gray}{\pm #2}$}

\DeclareMathOperator*{\argmin}{arg\,min}

\newcommand{\method}{\textsc{Polar}\xspace}

\title{Efficient Adaptive Data Acquisition via \\ Pretrained Belief Representations}

\author{
\textbf{Daolang Huang}$^{1,2}$ \quad
\textbf{Zhuoyue Huang}$^{5}$ \quad
\textbf{Conor Hassan}$^{1,2}$ \quad \\[1em]
\textbf{Luigi Acerbi}$^{3}$ \quad
\textbf{Samuel Kaski}$^{1,2,4}$ \quad 
\textbf{Tom Rainforth}$^{5}$ \quad \\[1em]
$^{1}$ ELLIS Institute Finland \quad
$^{2}$ Department of Computer Science, Aalto University, Finland \\[0.2em]
$^{3}$ Department of Computer Science, University of Helsinki, Finland \\[0.2em]
$^{4}$ Department of Computer Science, University of Manchester, UK \\[0.2em]
$^{5}$ Department of Statistics, University of Oxford, UK
}

\begin{document}

\maketitle

\begin{abstract}
    Learning effective policies for adaptive data acquisition remains challenging: methods that select queries using explicit posterior approximations can be limited by the difficulty of inference, while policies learned directly from observation histories must jointly learn useful representations and acquisition strategies, making learning harder. To address this, we introduce policy learning with belief representations (\method), based on the insight that optimal data acquisition depends on the observation history only through a sufficient belief state. Specifically, \method decouples representation learning from policy learning by leveraging pretrained predictive foundation models as belief-state encoders and training a policy head on top of their representations.
    This yields a simple, unified amortised policy learning framework for Bayesian experimental design, Bayesian optimisation, and active learning, differing only in the task-specific utility used to train the policy. Empirically, we find that \method outperforms state-of-the-art amortised methods across diverse tasks while requiring far fewer training samples, demonstrating a significant step in the scalability and efficiency of amortised data acquisition.
\end{abstract}

\vspace{-2mm}
\section{Introduction}
\vspace{-2mm}

Many sequential learning problems can be posed as \emph{adaptive data acquisition}: at each round, a learner chooses which query to make next to improve a downstream decision based on the data observed so far, receives the resulting observation, and repeats until a budget is exhausted. This abstraction covers Bayesian experimental design (BED) \citep{lindley1956measure,chaloner1995bayesian,rainforth2024modern}, Bayesian optimisation (BO) \citep{garnett2023bayesian}, and active learning (AL) \citep{settles2012active}. These settings differ in their downstream objectives, but share the same central challenge: \emph{mapping a growing observation history to acquisition decisions under a finite budget}.

\looseness=-1
Learning effective acquisition policies in this setting is difficult. Existing approaches fall into two families: those that work through an explicit approximation of the posterior, and those that map observation histories directly to decisions. Both are problematic in different ways. Posterior-based methods \citep{chang2025amortized,muller2023pfns4bo,li2025none} allow the model to be exploited in the policy architecture itself and thus efficiently identify what aspects of the data are most important or our task.
However, they must either perform expensive inference during the experiment or learn complex amortised inference networks, which is sometimes more challenging than the adaptive design problem itself~\citep{foster2021deep}, while errors in the inference can compound through the sequential acquisition process.
Mapping directly from observation histories directly to the next decision \citep{foster2021deep,ivanova2021implicit,huang2024amortized,maraval2023end} avoids explicit posterior inference, but pays for this by not exploiting model structure and invariances, which makes the policy more difficult to learn, they must implicitly learn how the posterior changes with the data. In other words, they must simultaneously learn both how to \emph{represent} the current dataset and how to \emph{act} on that representation.
The former can be as challenging as the latter, potentially becoming a bottleneck that increases the number of samples needed to train the policy.

\begin{figure}[t] 
    \centering
    \includegraphics[width=\linewidth]{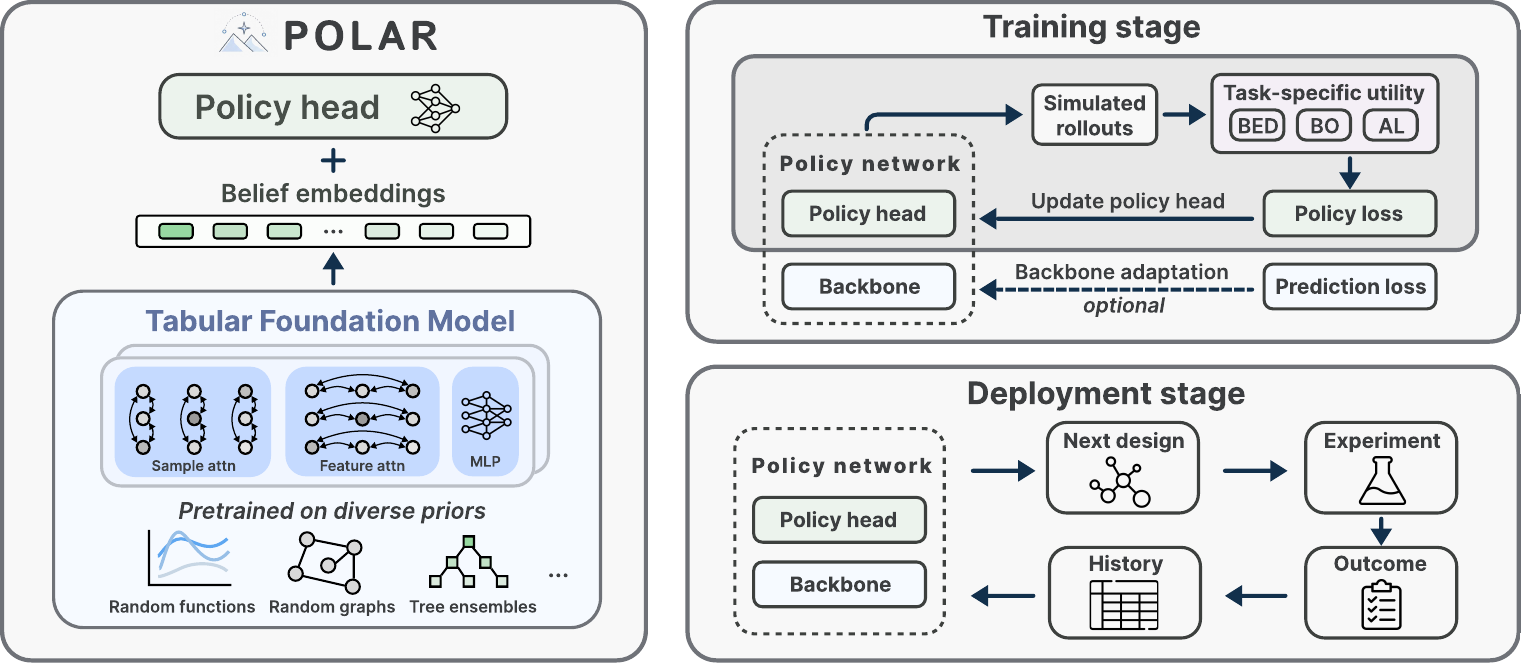}
    \vspace{-4mm}
    \caption{\textbf{Overview of \method.}  \emph{Left}: \method uses a pretrained tabular foundation model as a belief encoder and trains a policy head on top of it. \emph{Top right}: The policy loss updates only the policy head using task-specific acquisition utilities. An optional auxiliary prediction loss adapts the backbone through supervised prediction on additional target points. \emph{Bottom right}: At deployment, the policy maps the current history and the step to the next design in a single forward pass.}
    \vspace{-4mm}
    \label{fig:main_gist}
\end{figure}

\looseness=-1
Our key insight that allows us to address these limitations is that, while optimal sequential decisions depend only on the observation history through the corresponding posterior, a learned policy need not access this posterior explicitly: it may instead act on any \emph{representation} that preserves the information needed to recover the posterior.
This suggests a beneficial middle ground between raw observation histories and explicit posterior estimates that separates this representation learning from how to map representations to decisions. To exploit this insight, we therefore introduce a pre-trained intermediate \emph{belief representation} that captures underlying features in how the posterior varies with the data, without over-committing to a single posterior approximation. 
This then simplifies policy learning while retaining information that an explicit posterior approximation may discard.

\looseness=-1
Pretrained predictive foundation models provide a natural starting point for this belief representation. Namely, rather than learning a representation from scratch, we can build on models whose pretraining has already shaped them to organise observed data into useful representations for a broad family of tasks.
We instantiate this idea with Tabular Foundation Models (TFMs; \citealp[]{muller2025position}), such as TabPFN \citep{hollmann2025accurate,grinsztajn2025tabpfn} and TabICL \citep{qu2025tabicl,qu2026tabiclv2}, which produce candidate-conditioned representations from a context set in a single forward pass. 

Specifically, we introduce \textbf{PO}licy \textbf{L}e\textbf{A}rning with Belief \textbf{R}epresentations (\method), an amortised data acquisition framework that 
uses the hidden representations from a TFM as our belief representation and trains a policy head to select the next query from the encoding this provides. In doing so, \method shifts amortised acquisition from jointly learning representations and decisions to learning how to act on pretrained belief representations. An overview of \method is shown in \Cref{fig:main_gist}.

In summary, our main contributions are:\vspace{-2pt}
\begin{enumerate}[leftmargin=2em, topsep=0pt]
    \item We give a unified decision-theoretic view of adaptive data acquisition as learning acquisition policies on top of belief representations, rather than learning sequential decision-making directly from raw histories or acting through explicit posterior approximations.
    \vspace{-2pt}
    \item We introduce a simple decoupled policy architecture that uses pretrained predictive foundation models as belief-state encoders and trains a policy head on top.
    \vspace{-2pt}
    \item Across diverse adaptive data acquisition tasks, the learned policy substantially outperforms greedy acquisition on the same backbone and exceeds state-of-the-art amortised methods while requiring up to $100\times$ fewer task-specific training samples.
\end{enumerate}


\section{Preliminaries} \label{sec:background}

\textbf{Adaptive data acquisition.} We consider a finite-horizon adaptive data acquisition problem in which a learner repeatedly selects designs based on the data observed so far.  A latent world state $\theta \sim p(\theta)$ determines observations through a model $p(y\mid x, \theta)$, where $x\in\Xcal$ is a design and $y\in\Ycal$ is the corresponding outcome. At round $t$, the learner has observed $\Dcal_{t-1}=\{(x_i, y_i)\}_{i=1}^{t-1}$, chooses a feasible design $x_t \in \Xcal_t(\Dcal_{t-1}) \subseteq \Xcal$, observes $y_t \sim p_y(\cdot \mid x_t, \theta)$, and appends $(x_t, y_t)$ to the history. We aim to optimise an acquisition policy $\pi=(\pi_1,\ldots,\pi_T)$, consisting of a sequence of decision rules $\pi_t:\mathcal{H}_{t-1}\to \Xcal$ satisfying $\pi_t(\Dcal_{t-1})\in \Xcal_t(\Dcal_{t-1})$. Different adaptive data acquisition problems share this sequential structure, but differ in the downstream objective and which aspect of the world state matters for this. For example, in many BED problems, we seek information about unknown system parameters $\theta$ \citep{chaloner1995bayesian,ryan2016review,rainforth2024modern}; in active learning, we want to learn to make effective future predictions \citep{smith2023prediction}; and in BO, we seek evaluations that rapidly identify a global optimum \citep{garnett2023bayesian}.

\looseness=-1
Existing approaches often choose each design by acting through an explicit approximation of the posterior: given $\Dcal_{t-1}$, they approximate the posterior, 
either by directly performing inference on the underlying model or using an amortised neural predictor \citep{chang2025amortized, muller2023pfns4bo, li2025none}, and then select the next design by optimising an acquisition function. Examples include the traditional constructions of BO, AL, and BED, which iterate between updating the model and explicitly maximising an expected utility, such as the Expected Improvement (EI) in BO \citep{garnett2023bayesian} or the Expected Information Gain (EIG) in BED \citep{lindley1956measure, lindley1972bayesian}.
These methods rely on the quality of the posterior approximation exposed to the acquisition rule. Outside rare cases where exact inference is tractable, posterior approximation is hard, and its errors can propagate into suboptimal acquisitions  \citep{lacoste2011approximate,maus2024approximation}, especially across multiple rounds.

An alternative is to learn a policy that maps observation histories directly to designs.
In this case, a parameterised policy is trained from simulated trajectories to maximise a task-specific sequential utility, so that test-time decisions require only a forward pass rather than the optimisation of a hand-designed acquisition function. For example, amortised BED methods \citep{foster2021deep, ivanova2021implicit, blau2022optimizing,blau2023statistically,huang2024amortized,huang2025aline,bracher2025jadai, phillips2026bayesian} map observation histories directly to experimental designs, and related end-to-end policies have appeared for BO \citep{maraval2023end, zhang2025context} and AL \citep{huang2025aline, li2025amortized}. This amortises test-time decision-making, but requires the same model to learn both a representation of the current dataset and an acquisition strategy from sequential supervision alone.
  
\textbf{Pretrained predictive models.} Recent work on in-context learning has produced models that can approximate predictive distributions in a single forward pass, without iterative fitting or gradient updates. Transformer Neural Processes \citep{nguyen2022transformer} and Prior-Fitted Networks \citep{muller2021transformers} learn to map a context set of observed input-output pairs to predictive distributions for new inputs, after offline training on large families of synthetic tasks. More recent Tabular Foundation Models (TFMs), such as TabPFN \citep{hollmann2025accurate,grinsztajn2025tabpfn} and TabICL \citep{qu2025tabicl,qu2026tabiclv2}, scale this idea to practical tabular prediction, achieving strong performance across a wide range of supervised learning problems. Given a dataset $\Dcal$ and a candidate input $x$, these models produce a predictive distribution $q_\psi(\cdot \mid x, \Dcal)$ for the corresponding outcome, together with internal representations $r_\psi(x,\Dcal)$ that depend on both $x$ and $\Dcal$. These representations are not designed for any particular sequential task, but can encode predictive uncertainty, local structure, and context-dependent function behaviour. 

\section{A Unified Decision-Theoretic View of Adaptive Data Acquisition} \label{sec:insight}

To understand policy learning for adaptive data acquisition, we must first formalise what an ideal representation should capture. We do so by casting adaptive data acquisition under a unified Bayesian decision-theoretic framework.

\textbf{Bayes-optimal policy.} Consider a finite-horizon sequential data acquisition problem. Once all data has been acquired, a downstream action, $a \in \mathcal{A}$, is chosen, which incurs loss $\ell(a, z)$, where $z=\tau(\theta)$ represents the aspects of the overall world state that directly impact our loss. Depending on context, $a$ might be an estimate, a predictive distribution, or a physical decision.
Following Bayesian decision theory \citep{savage1951theory,lindley1972bayesian,berger1985statistical,huang2026loss} and defining $p(z \mid \Dcal_T)$ as the pushforward of our posterior onto $z$, the Bayes-optimal acquisition policy is now:
\begin{equation}
\pi^\star =
\argmin_{\pi \in \Pi}
\mathbb E_{p(\Dcal_T;\pi)}
\!\left[
\min_{a \in \mathcal A}
\mathbb E_{p(z \mid \Dcal_T)}[\ell(a,z)]
\right].
\label{eq:general_design_obj}
\end{equation}
Here the expected loss of the Bayes action, $\min_{a \in \mathcal A}
\mathbb E_{p(z \mid \Dcal_T)}[\ell(a,z)]$, is a function only of our posterior beliefs for a given loss function.
We can thus rewrite~\Cref{eq:general_design_obj} as an expected functional of the posterior \citep{rossa2026action}: defining $s(q,z) := \ell(a^\star(q), z)$ where $a^\star(q):= \arg\min_{a \in \mathcal{A}} \mathbb{E}_{z \sim q}[\ell(a,z)]$, then,
\begin{equation}
    \pi^\star 
    =\argmin_{\pi \in \Pi}
\mathbb E_{p(\Dcal_T;\pi)p(z|\Dcal_T)}
\!\left[
s\!\left(p_z(\cdot \mid \Dcal_T), z\right)
\right]
= \argmin_{\pi \in \Pi}
\mathbb E_{p(\theta)p(\Dcal_T|\theta;\pi)}
\!\left[
s\!\left(p_z(\cdot \mid \Dcal_T), z\right)
\right].
\label{eq:posterior_score_w}
\end{equation}
It turns out that this $s$ is always a proper scoring rule for distributions on $z$~\citep{savage1971elicitation,gneiting2007strictly}.
We can thus view~\Cref{eq:posterior_score_w} as an \emph{expected posterior uncertainty}, $\mathbb{E}_{p(\Dcal_T;\pi)}[h_s[p_z(\cdot \mid \Dcal_T)]]$, where our uncertainty measure is the \emph{generalised entropy} $h_s[q]=\mathbb{E}_{z\sim q}[s(q,z)]$~\citep{dawid1998coherent,bickfordsmith2025}.


Under this unified foundational framework of sequential Bayesian decision-making, we can thus equivalently think about defining adaptive data gathering problems through defining either downstream losses, scoring rules, or uncertainty measures, all of which also (implicitly) define the target variables of interest for our data gathering, $z$.
Most existing BED and active learning methods are explicitly based on defining uncertainty measures, which is typically taken to be entropy to yield the expected information gain~\citep{lindley1956measure,rainforth2024modern}, though they can vary in what variables are targeted for learning, e.g., in AL, we can either take $z$ to be model parameters~\citep{houlsby2011bayesian} or downstream predictions~\citep{smith2023prediction}.
In BO, it is more common to work directly with a downstream loss, with $z$ taken to be the true function values at the queried inputs.  For example, we can recover a strategy equivalent to targeting expected improvement by taking $\ell_{\mathrm{EI}}(a,f)=-f(a)$ and then $\mathcal{A}$ as the set of inputs we have queried (such that the minimisation over $a$ trivially yields the point with the highest posterior mean).

\textbf{From posterior beliefs to belief representations.} 
We can further rearrange \Cref{eq:posterior_score_w} to make clear the dependency of the optimal policy on historical data as follows (where $\Dcal_{t:T} = \{(x_i,y_i)\}_{i=t}^{T}$)
\begin{align*}
    \pi^\star =\argmin_{\pi \in \Pi}
\mathbb E_{p(\Dcal_{t-1} ; \pi)} \left[ \mathbb{E}_{p(\theta | \Dcal_{t-1} ) p(\Dcal_{t:T} | \theta ; \pi)}
\!\left[
s\!\left(
p_{z}(\cdot | \Dcal_T),
z\right) 
\right]\right],
\end{align*}
where because $p(z | \Dcal_T)$ is a pushforward of $p(\theta | \Dcal_T)$, it only depends on $\Dcal_{t-1}$ through the intermediate world state posterior $p(\theta|\Dcal_{t-1})$.
Once $\Dcal_{t-1}$ has been observed, the outer expectation disappears for the remaining decisions as $\Dcal_{t-1}$ is fixed.
This now yields a crucial insight: 
the Bayes-optimal policy depends on the history only through the current world state posterior $p(\theta|\Dcal_{t-1})$ and the current step $t$ (which is required for knowing how many steps are left in our process, as this affects things like exploration-exploitation trade-offs).
Equivalently, 
there exist maps $f_{\mathcal{D}},\,f_{\mathrm{post}}$ such that
\begin{equation}
\pi_t^\star(\mathcal{D}_{t-1})
\;=\; f_{\mathcal{D}}(\mathcal{D}_{t-1})
\;=\; f_{\mathrm{post}}\!\left(p_\theta(\cdot \mid\mathcal{D}_{t-1}), t\right).
\label{eq:two_equivalent_forms}
\end{equation}
Thus, any two datasets that induce the same world state posterior at the same step should lead to the same optimal next action. This invariance suggests that the right input to policy learning is not necessarily the raw dataset itself, but some representation that preserves the information in $\mathcal{D}_{t-1}$ needed to recover our world state belief.
In particular, we want a representation that faithfully retains this information, while also providing the simplest possible mapping to the optimal policy decisions.

To this end, we form $e_t=(\mathrm{enc}(D_{t-1}),t)$ using a belief encoder $\mathrm{enc}$ and the current step. We call $e_t$ a belief representation if there exists a decoder $\mathrm{dec}$ such that
\begin{equation}
\operatorname{dec}(e_t)
= \left(p_\theta(\cdot\mid \Dcal_{t-1}),t\right).
\label{eq:decoder}
\end{equation}
Whenever the decoding condition above holds, there also exists a policy head $g$ for which
\begin{equation}
    \pi_t^\star(\mathcal{D}_{t-1}) = g(e_t),
    \label{eq:h_t}
\end{equation}
because the posterior and current step are determined by $e_t$. Two extreme choices of $e_t$ recover the two dominant approaches in the literature:
\begin{itemize}[leftmargin=1.5em]
    \vspace{-2mm}
    \looseness=-1
    \item \textbf{Raw history: $e_t = \mathcal{D}_{t-1}$}. This is trivially a belief representation, since the posterior can be recovered from the history and the current step is implied by the dataset itself. This is the input used by direct policy-learning methods, but it leaves the policy head $g$ with the entire job of capturing the dependency of the posterior on the data, significantly complicating practical training.
    \item \looseness=-1 \textbf{Posterior: $e_t = (p_\theta(\cdot \mid\mathcal{D}_{t-1}), t)$}. This is the belief state which yields a trivial decoder (the identity function) and thus incorporates all desired posterior invariances (though there may be further invariances in the mapping from intermediary posteriors to expected future uncertainties), thus making the training of $g$ much easier.
    In practice, however, this belief state cannot usually be calculated exactly and is replaced by an approximation $\hat p_\theta(\cdot\mid\mathcal{D}_{t-1})$, 
    no longer guaranteeing the decoding condition in \Cref{eq:decoder} and exposing $g$ to the approximation error of the inference step.
\end{itemize}

\vspace{-1mm}
Rather than going with either of these extremes, our key idea is to use an encoder that avoids needing an overly complex $g$ while still retaining information useful for acquisition, without restricting the policy to an explicit posterior approximation. 
We achieve this by using encodings that provide effective \emph{representations} of the posterior, without over-committing to using precise posterior approximation.
\Cref{sec:method} shows how predictive foundation models provide an appropriate and practical encoder as a by-product of their training.




\section{Policy Learning with Belief Representations}
  \label{sec:method}

Section~\ref{sec:insight} argued that acquisition policies should act on representations of the world state posterior and the current step: representations that retain the information in observation histories needed for optimal decisions, while structured enough to make the map to acquisitions easy to learn. 
Pretrained predictive foundation models are natural candidates for this role: to predict well across diverse tasks from in-context observations alone, such models are encouraged to summarise the observed data into representations that track how beliefs change as data accumulates, which is useful for acquisition.
We now describe how \method instantiates this principle using Tabular Foundation Models (TFMs) as practical belief encoders.

\textbf{Foundation models as belief encoders.} 
A pretrained TFM $F_\psi$ is trained to map context sets to predictive distributions across a broad family of synthetic tasks \citep{hollmann2025accurate,qu2025tabicl}. 
Given a dataset $\mathcal{D}_{t-1}$ and a candidate $x\in\mathcal{X}$, a forward pass returns (i)~a predictive distribution $q_\psi(\cdot\mid x,\mathcal{D}_{t-1})$ over the outcome $y$, and (ii)~a candidate-conditioned hidden representation $r_\psi(x,\mathcal{D}_{t-1})\in\mathbb{R}^{d_r}$. The predictive distribution $q_\psi$ is decoded from $r_\psi$ through a fixed prediction head. It lives in outcome space, approximating a posterior predictive over $y$ at a candidate $x$, whereas the decision-theoretic object in \Cref{sec:insight} is the posterior over the world state $\theta$. Thus, although $q_\psi$ is useful for prediction and for some task-specific surrogates, it is not in general itself a belief representation for $\theta$. 


We therefore use the hidden representation $r_\psi$, rather than $q_\psi$, as the interface for policy learning. The predictive head may discard useful features when mapping $r_\psi$ to $q_\psi$. Training a shared model to support predictions at arbitrary inputs may further encourage features that capture how the data constrain the underlying predictive function. These considerations do not, however, guarantee that $r_\psi$ is sufficient for the posterior over $\theta$. For a given history, even exact marginal predictions at every input do not determine the joint predictive distribution over future observations. The importance of this distinction depends on the acquisition objective: marginal predictions suffice for some myopic criteria, while joint dependencies can matter for acquisition decisions, especially when planning ahead. Motivated by \Cref{sec:insight}, we therefore pair $r_\psi$ with the current step $t$ as an approximate belief representation and assess its usefulness empirically in \Cref{sec:exp}.

While the framework of \Cref{sec:insight} applies to general design spaces, TFMs are far more efficient when the context $\mathcal{D}_{t-1}$ is processed once and reused across a finite candidate set. We therefore work in a pool-based setting: at each round, the policy chooses from a feasible pool $\mathcal{C}_t \subseteq \mathcal{C}$. A single batched pass over $\mathcal{C}_t$ returns the candidate-conditioned representations $\{r_\psi(x,\mathcal{D}_{t-1}) : x\in\mathcal{C}_t\}$. This setting matches many practical problems: hyperparameter optimisation over discrete configuration grids, molecular optimisation over finite molecule libraries, and active learning over finite unlabelled pools. For continuous design spaces, the same architecture can be combined with a candidate generator or evaluated over sampled candidate sets.



\textbf{Policy head.}
We instantiate the policy head $g$ as a function $g_\eta:\mathbb{R}^{d_r+1}\to\mathbb{R}$ with learnable parameters $\eta$. For each candidate $x\in\mathcal{C}_t$, $g_\eta$ maps $r_\psi(x,\mathcal{D}_{t-1})$, concatenated with the number of remaining acquisitions $T-t+1$ normalised by the training horizon, to a scalar logit, and normalising these logits across the candidate pool with a softmax gives the stochastic policy $\pi_\eta(\cdot\mid\mathcal{D}_{t-1},\mathcal{C}_t, t)$.
We parameterise $\pi_\eta$ as a distribution over candidates during training, which provides exploration and admits standard stochastic-gradient policy training. 
At deployment, we select the candidate with the largest learned score.

\textbf{Task scoring rules.}
The architecture is shared across all tasks; only the task-specific scoring rule changes. 
For each task, we define $\widehat{\Phi}(\mathcal{D}_T,z)$ as a tractable surrogate of $-s(p_z(\cdot | \Dcal_T),z)$. 
In some tasks, this score is directly observable, while in others a surrogate estimator is needed. In BED, the natural scoring rule is the log loss on a target variable, whose expected reduction is the EIG. Since the EIG is intractable, we approximate it with the sPCE lower bound \citep{foster2021deep}. Other EIG estimators or losses could be substituted without further change. In BO, the queried function values $y_t$ are directly observed, so the score reduces to a tractable function of the trajectory. We use the terminal best observed value,
$\widehat{\Phi}_{\mathrm{BO}}(\mathcal{D}_T,z)=\max_{t\leq T} y_t$,
for which no posterior approximation is required. In AL, when the goal is to make future predictions \citep{smith2023prediction}, the terminal score is the predictive log-density on a target set, which is also intractable, so we use a variational surrogate built on $q_\psi$.


\textbf{Policy training.}
We train the policy using reinforcement learning on simulated tasks to maximise
the expected terminal score
$\mathbb{E}_{p(\theta)p(\mathcal D_T\mid\theta;\pi_\eta)}
[\widehat\Phi(\mathcal D_T,z)]$. Since the policy selects among discrete candidates, we use score-function gradients rather than reparameterisation; these also do not require differentiating through the simulator. To provide a dense training signal, we express the terminal objective through per-step utilities $u_t=\widehat\Phi(\mathcal D_t,z)-\widehat\Phi(\mathcal D_{t-1},z)$, so that $\sum_{t=1}^T u_t =\widehat\Phi(\mathcal D_T,z)-\widehat\Phi(\mathcal D_0,z)$. For each sampled trajectory, we use the policy-gradient surrogate
\begin{equation}
\mathcal L_{\mathrm{pol}}(\eta) = -\sum_{t=1}^T G_t^{(\gamma)} \log\pi_\eta(x_t\mid\mathcal D_{t-1},\mathcal C_t, t),
\qquad
G_t^{(\gamma)} = \sum_{s=t}^T \gamma^{s-t}u_s,
\label{eq:policy_loss}
\end{equation}
where $\gamma\in[0,1]$. We average this loss over sampled trajectories, treating the sampled data and $G_t^{(\gamma)}$ as fixed when differentiating with respect to $\eta$.



With $\gamma=1$, the update includes all subsequent utility increments, as required by the terminal-score objective. Smaller values of $\gamma$ provide a training approximation: in particular, $\gamma=0$ uses only the immediate utility $u_t$, giving a myopic training signal. This can simplify learning and reduce gradient variance, but ignores an action's effects on later acquisitions. We use $\gamma=0$ in our main experiments, where this approximation is empirically effective. Appendix~\ref{app:pendulum} additionally considers a dynamical task where planning ahead is important and $\gamma=1$ substantially improves performance.

\textbf{Backbone adaptation.}
The pretrained TFM backbone $F_\psi$ is trained on synthetic priors that may not perfectly match the downstream task distribution. We therefore optionally adapt $\psi$ using an auxiliary supervised prediction loss,
$\mathcal{L}_{\mathrm{pred}}(\psi)=\sum_{t=1}^T\sum_{m=1}^M\ell_{\mathrm{pred}}\!\big(q_\psi(\cdot\mid x_m^\star,\mathcal{D}_{t-1}),y_m^\star\big)$,
where $\{(x_m^\star,y_m^\star)\}_{m=1}^M$ are target points sampled from the same simulated task during training. The form of $\mathcal{L}_{\mathrm{pred}}$ depends on the prediction head: for density-based heads \citep{hollmanntabpfn}, it is negative log-likelihood, while for quantile-based heads \citep{qu2026tabiclv2}, it is quantile regression loss. A key design choice is that policy gradients are not propagated into the pretrained backbone: $\mathcal{L}_{\mathrm{pol}}$ updates only the policy head, while $\mathcal{L}_{\mathrm{pred}}$ is the only loss that may update $\psi$. This adapts the backbone through supervised prediction while preventing high-variance policy gradients from destabilising it, keeping representation refinement and decision learning decoupled.

\textbf{Limitations and possible extensions.} We close this section by noting the scope of the particular realisation of \method described above, and the natural extensions it suggests. 
First, our policy acts over finite candidate pools. This covers the pool-based settings considered in this work, but problems requiring highly precise continuous designs may benefit from continuous-action variants, for example, by optimising the learned score over generated candidates or by introducing a global summary token that feeds a continuous policy head. 
Second, distribution shift can affect \method at two stages. A mismatch between TFM pretraining and downstream policy training may degrade the learned representation and thereby the decision rule learned by the downstream policy. In particular, histories requiring different acquisition decisions may be mapped to similar representations, leaving the policy head unable to recover the lost distinction. The auxiliary prediction loss can realign the representation, while designing more transferable TFM pretraining priors is a complementary direction \citep{zhang2026mitra, bouadi2026shaping}. A mismatch between policy training and deployment can instead render the learned policy misspecified, even when the representation remains informative. Broader or deployment-matched training distributions and misspecification-robust policy-learning methods \citep{tang2025generalization,forster2025improving} are promising remedies.

\vspace{-1mm}
\section{Related Work}
\vspace{-1mm}

\looseness=-1
Early policy-based BED used approximate dynamic programming with online design optimisation \citep{huan2016sequential}. Amortised policy-based BED was subsequently proposed by DAD \citep{foster2021deep}, which learns a design policy offline and deploys it in a single forward pass, with iDAD \citep{ivanova2021implicit} extending the framework to implicit likelihood models. A parallel line recasts the problem as reinforcement learning \citep{blau2022optimizing,lim2022policy}. \citet{hedman2025step} introduce a semi-amortised variant that combines an offline policy with online refinement, while \citet{guo2026constrained} consider constraint-aware BED via online planning, and a recent thread improves training signals through alternative EIG estimators \citep{blau2023statistically,iqbal2024nesting,shen2025variational,bracher2025jadai}. 
ScoreBED \citep{phillips2026bayesian} uses approximate marginal-likelihood scores for policy learning. Related decision-theoretic formulations target downstream losses \citep{huang2024amortized, huang2026loss, rossa2026action}.
\citet{huang2025aline} show that prediction-driven supervision can support policy learning, but its shared backbone is trained from scratch, making it sample-inefficient.
Our work is complementary to all of these; we reuse a pretrained predictive model as the belief encoder so that a small policy head can be trained with standard policy gradient.

\looseness=-1
In BO, one line of work retains the classical surrogate-plus-acquisition template while replacing the surrogate with increasingly powerful neural predictors, including PFN- and transformer-based models \citep{muller2023pfns4bo,chang2025amortized,li2025none,hungboformer,viering2025alpha}. A second line instead learns the acquisition policy end-to-end, mapping observed trajectories directly to the next query \citep{chen2017learning,chen2022towards,maraval2023end,song2024reinforced,zhang2025context,zhangpabbo}. \method sits between these threads: it inherits the strong representations of a pretrained surrogate, but learns a policy on top of those representations rather than committing to a fixed acquisition rule.

Our work is also related to the broader family of meta-learned predictors that map context sets to predictive distributions. Originating from the Conditional Neural Processes \citep{garnelo2018conditional}, these architectures evolved into Transformer Neural Processes \citep{nguyen2022transformer} and early PFNs \citep{muller2021transformers}. Recent tabular foundation models such as TabPFN \citep{hollmanntabpfn,hollmann2025accurate,grinsztajn2025tabpfn} and TabICL \citep{qu2025tabicl,qu2026tabiclv2} scale this paradigm to broad supervised-learning regimes and have stimulated a growing literature \citep{muller2025position,ye2025closer,hoo2024tabular,zhang2025tabpfn}. Finally, since our backbone is adapted with an auxiliary prediction loss, our work is also related to recent studies on fine-tuning TFMs to better match downstream data distributions \citep{rubachev2025finetuning,tanna2025tabtune,garg2025real}.

\section{Experiments} \label{sec:exp}

We now empirically evaluate \method across a range of tasks. \Cref{sec:bed_exp} benchmarks \method on two standard BED tasks, location finding and constant elasticity of substitution, where we also report a series of ablations isolating the contribution of each component of our method. \Cref{sec:hpo} evaluates \method for BO on a hyperparameter optimisation benchmark, and \Cref{sec:dockstring} turns to a real-world high-dimensional molecular optimisation task. We additionally study loss-driven active learning in Appendix~\ref{app:al_exp}, which demonstrates that \method extends naturally to settings defined by user-specified predictive losses. Further experimental results and ablation studies are provided in Appendix~\ref{app:additional_exps}. Unless otherwise stated, all experiments use TabICL v2 \citep{qu2026tabiclv2} as the backbone; additional results with TabPFN v2.5 \citep{grinsztajn2025tabpfn} are reported in Appendix~\ref{app:location_finding_exp}.
The code to reproduce our experiments is available at: \url{https://github.com/huangdaolang/POLAR}.

\subsection{Benchmarking on Bayesian experimental design tasks} \label{sec:bed_exp}
We begin with two standard BED benchmarks, location finding and constant elasticity of substitution (CES), which have been used extensively in prior work \citep{foster2019variational,foster2020unified,blau2022optimizing,huang2025aline}. The goal of location finding \citep{sheng2005maximum} is to infer the locations of $K$ hidden sources in $d_x$-dimensional space from noisy distance measurements. We consider two variants to test scalability: a standard 2D setting ($d_x=2, K=2, T=30$) and a higher-dimensional 5D setting ($d_x=5, K=2, T=50$). CES \citep{arrow1961capital} involves eliciting the parameters of an economic utility function from a series of pairwise preference queries.
We compare against Random, DAD \citep{foster2021deep}, RL-BOED \citep{blau2022optimizing}, and ALINE \citep{huang2025aline}. For both tasks, we report the EIG lower bound estimated by sPCE, and the upper bound estimated by sNMC \citep{foster2021deep}.
Full details on the task and training setups are provided in Appendix~\ref{app:bed}.

\begin{figure}[t]
  \centering
    \subfigure[]{\includegraphics[width=0.31\linewidth]{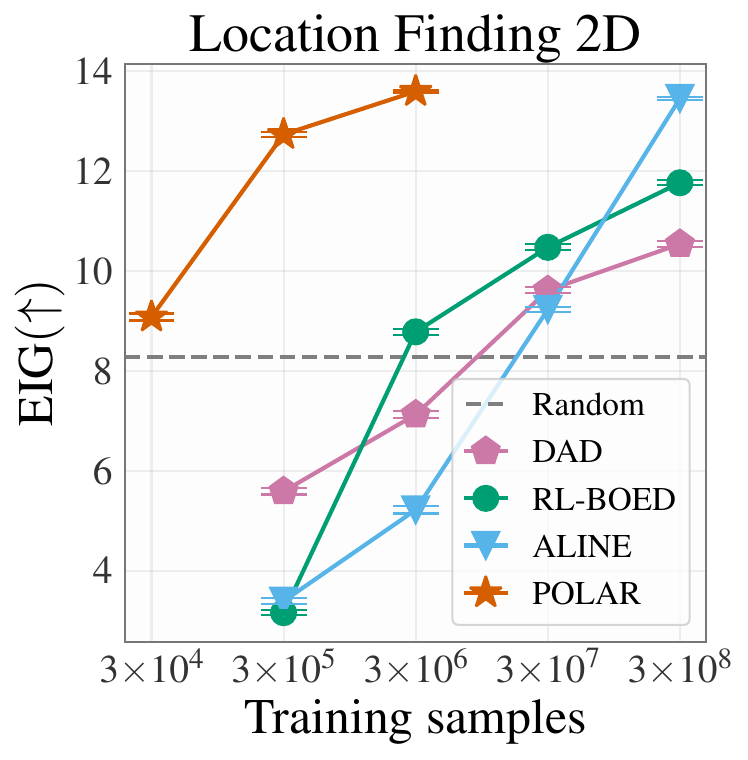}}
    \subfigure[]{\includegraphics[width=0.31\linewidth]{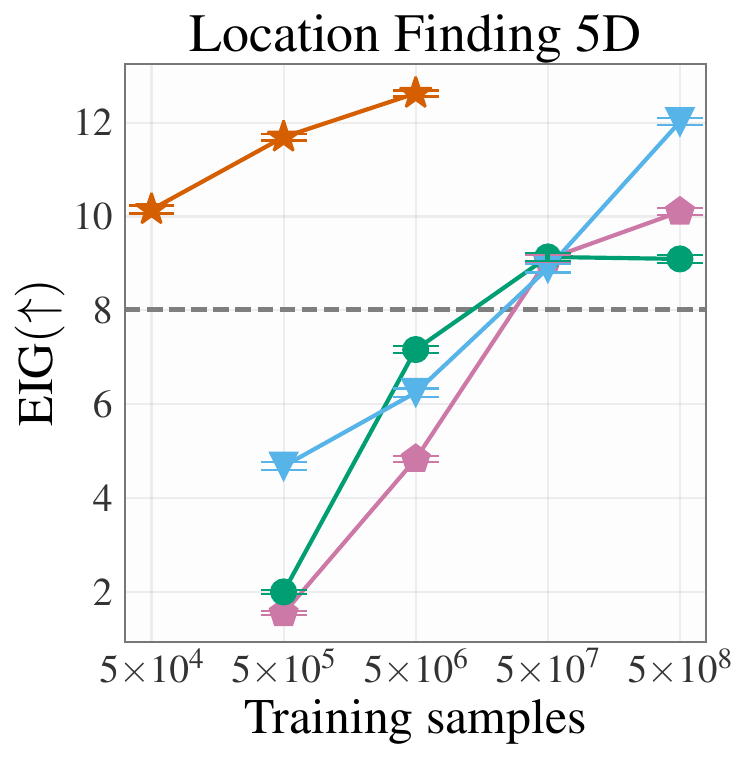}}
    \subfigure[]{\includegraphics[width=0.36\linewidth]{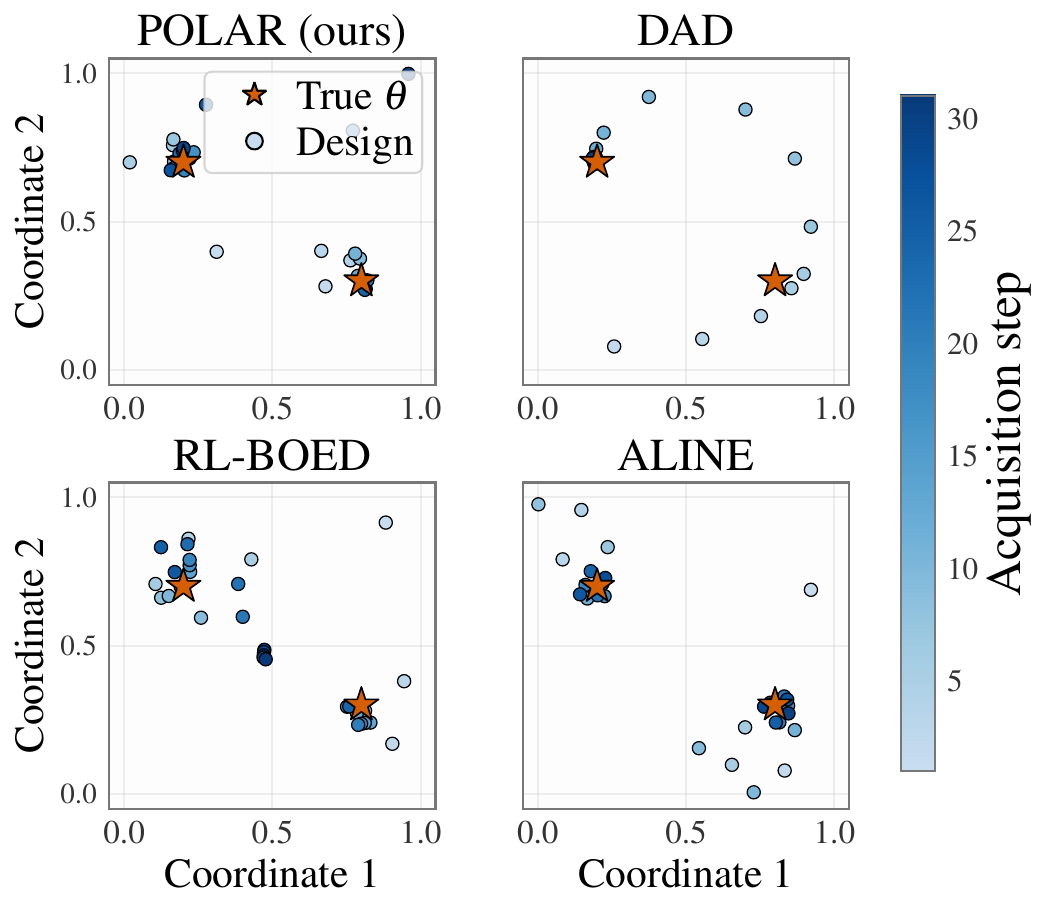}}
  \caption{\textbf{Location finding}. (a) EIG against total training samples in the 2D setting. Error bars denote standard error across 1000 runs. (b) The same comparison in the 5D setting.  (c) Example design trajectories for all methods on a shared latent source configuration.}
  \label{fig:location_finding}
\end{figure}

\paragraph{Main results.} To explicitly assess the training efficiency, we report performance as a function of the total number of training samples (see Appendix~\ref{app:bed} for details on how we calculate this metric for a fair comparison). For location finding, Figures \ref{fig:location_finding}a and \ref{fig:location_finding}b show that \method is dramatically more sample-efficient than prior amortised BED policies. In the standard 2D task, our method slightly outperforms ALINE, the current state-of-the-art, using approximately 100$\times$ fewer training samples, and surpasses the best reported results of both RL-BOED and DAD using 1,000$\times$ fewer samples. This dramatic efficiency gap further widens in the more complex 5D environment, where our method eclipses the peak performance of RL-BOED and DAD with 10,000$\times$ fewer samples, and exceeds ALINE with 100$\times$ fewer samples. 
This sample efficiency is particularly valuable in settings where simulation or real data collection is expensive, as is typical in scientific applications of experimental design. 
\Cref{fig:location_finding}c provides visualisations of design trajectories. The sNMC upper bound results are provided in Appendix~\ref{app:location_finding_exp}.
The conclusion remains consistent for the CES task. \Cref{tab:ces} shows that \method outperforms all amortised baselines at matched training-sample budgets. Results at full convergence for each method are reported in Appendix~\ref{app:ces_exp}, where \method remains competitive with the strongest baseline. Owing to the cost of evaluating a foundation-model backbone, our method is slightly slower to deploy than the other policies, but the absolute overhead remains negligible in practice: at roughly 0.03 seconds per decision, the latency is still effectively imperceptible at the human timescale. 

\begin{table}[t]
\centering
\caption{\textbf{CES}. Comparison of EIG estimates and deployment efficiency on the CES task. Results are reported as mean $\pm$ s.e. over 1,000 independent runs.}
\label{tab:ces}
\begin{tabular}{lccccc}
\toprule
\multirow{2}{*}{Methods} & \multicolumn{2}{c}{$10^5$ training samples} & \multicolumn{2}{c}{$10^6$ training samples} & \multirow{2}{*}{\makecell{Deployment \\ time (s)}}\\
\cmidrule(lr){2-3} \cmidrule(lr){4-5}
 & sPCE & sNMC & sPCE & sNMC & \\
\midrule
DAD  & \vpmse{9.54}{0.14} & \vpmse{9.66}{0.14} & \vpmse{11.46}{0.16} & \vpmse{12.33}{0.19} & \vpmse{0.0003}{0.00} \\
RL-BOED & \vpmse{11.04}{0.11} & \vpmse{11.21}{0.12} & \vpmse{12.32}{0.13} & \vpmse{14.73}{1.25} & \vpmse{0.0005}{0.00} \\
ALINE &  \vpmse{8.10}{0.18} & \vpmse{9.86}{0.29} & \vpmse{9.11}{0.17} & \vpmse{11.64}{0.33} & \vpmse{0.004}{0.00}\\
\method  & \vpmse{\textbf{11.15}}{0.15} & \vpmse{\textbf{12.09}}{0.19} & \vpmse{\textbf{13.10}}{0.12} & \vpmse{\textbf{16.24}}{0.25} & \vpmse{0.03}{0.00} \\
\bottomrule
\end{tabular}
\end{table}

\begin{wrapfigure}{r}{0.3\textwidth}
\vspace{-3mm}
    \begin{center}
        \includegraphics[width=\linewidth]{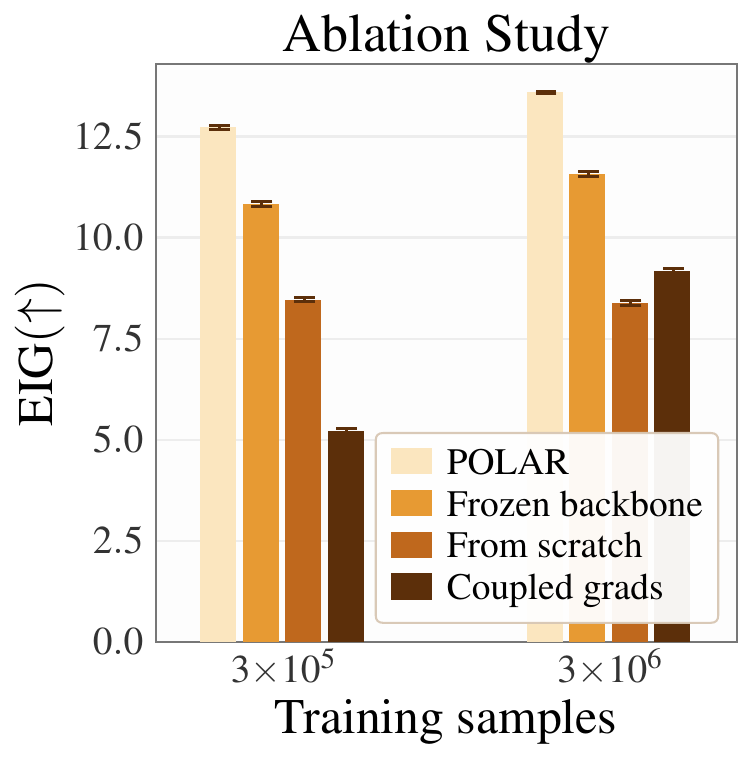}
    \end{center}
    \vspace{-2mm}
    \caption{Ablations for 2D location finding, isolating the impact of our core architectural choices, including gradient decoupling, backbone finetuning and initialisation.} 
    \label{fig:ablation}
\vspace{-4mm}
\end{wrapfigure}

To better understand where these gains come from, we conduct three ablations on the 2D location finding task: a frozen backbone, training from scratch, and coupled policy gradients (\Cref{fig:ablation}). 

\textbf{Frozen vs. finetuned backbone.} We first compare our default setting, in which the backbone is adapted by the supervised prediction loss, against a frozen-backbone variant. Even without any weight updates, the frozen variant remains competitive with the peak performance of RL-BOED. However, it clearly falls short of our finetuned variant. Because the backbone is a general-purpose foundation model pretrained on a vast array of synthetic distributions, its representations are not tailored to a specific task. The auxiliary prediction loss acts as an alignment signal, finetuning the backbone to adapt its predictive representations to the target domain, which in turn provides a better belief state for the policy head to act upon. 

\textbf{Pretrained backbone vs. training from scratch.} We next compare our method against the same architecture trained from scratch, without loading the pretrained checkpoint. The difference is substantial: training from scratch converges much more slowly and remains clearly worse within the same sample budget. This supports a central message of the paper: learning underlying belief representations is a critical part of the overall policy training for amortised adaptive design methods.
A pretrained foundation model has already significantly alleviated this problem, and starting from it confers the bulk of our sample efficiency. Additional ablations (Appendix~\ref{app:random_backbone}) show that the same policy-head architecture performs substantially worse with randomly initialised, frozen backbones, further supporting that our gains rely on pretrained representations and cannot be explained by the policy head alone.

\textbf{Decoupled vs. coupled policy gradients.} Our default method updates the backbone only through the prediction loss. We compare this against a coupled variant in which policy gradients are also allowed to update the backbone. The coupled variant severely degrades performance. Once policy gradients are allowed to modify the backbone, the same parameters must simultaneously satisfy two rather different objectives: a dense, relatively stable prediction objective and a high-variance reinforcement-learning objective. In practice, this appears to destabilise the learning and weaken the benefits of pretraining. By decoupling the gradients, we effectively assign the optimal learning objective to each component: low-variance supervised learning for belief state estimation, and policy gradients for decision-making on a stable representation manifold.

Additional ablations on backbone choice and representation layer are reported in \Cref{fig:location_finding_additional_ablations}, which shows \method is not tied to a single backbone, and later backbone layers provide more useful belief states for acquisition. Beyond acquisition performance, we also test whether the representations retain posterior information (Appendix~\ref{app:location_finding_exp}). A posterior decoder trained on the TFM embeddings achieves similar performance to one trained on raw observation histories.

\subsection{Hyperparameter optimisation benchmarks} \label{sec:hpo}

We next evaluate our method on BO using HPO-B \citep{arango2hpo}, a large-scale hyperparameter optimisation benchmark containing the evaluations of a wide range of machine learning models across thousands of hyperparameter configurations and datasets. HPO-B is pre-partitioned into multiple search spaces, each corresponding to a model family with its own hyperparameter parameterisation. Following prior work \citep{maraval2023end, li2025none}, we report results on six search spaces, spanning input dimensionalities from 2 to 16 and a representative range of model families. For each search space, we train our policy on the provided meta-training tasks and evaluate on the fixed test tasks and seeds defined by the benchmark.

We compare against baselines spanning both classical and amortised BO. \textsc{Random} samples configurations uniformly from the candidate pool. \textsc{GP} is a non-amortised baseline which fits a Gaussian process \citep{rasmussen2003gaussian} to the current observations. \textsc{Meta-GP}, following \citet{maraval2023end}, pretrains the kernel hyperparameters on the meta-training datasets and initialises the GP with these parameters at test time. \textsc{PFNs4BO} \citep{muller2023pfns4bo} uses a prior-fitted network \citep{muller2021transformers} for amortised prediction. \textsc{NAP} \citep{maraval2023end} learns an amortised acquisition policy using transformer neural processes \citep{nguyen2022transformer}. Finally, \textsc{TabICL} uses the same finetuned TabICL backbone as an amortised surrogate, but pairs it with a hand-designed acquisition function. All surrogate-based baselines use myopic EI as the acquisition strategy. Additional details on baseline configurations and the dataset are provided in Appendix~\ref{app:hpo}.

\paragraph{Results.} \Cref{fig:hpo} shows the average normalised regret and average rank aggregated across all search spaces. Our method achieves the lowest regret and the best average rank throughout the acquisition trajectory, with the gap to the strongest baselines opening within the first ten queries and persisting through the full budget. 
Notably, compared with \textsc{NAP}, our method shows that using pretrained TFMs can yield stronger performance without requiring the policy to learn its own belief representation from scratch. Additional per-search-space regret and rank curves are reported in Appendix~\ref{app:hpo_exp}.

\begin{figure}[t] 
    \centering
    \includegraphics[width=\linewidth]{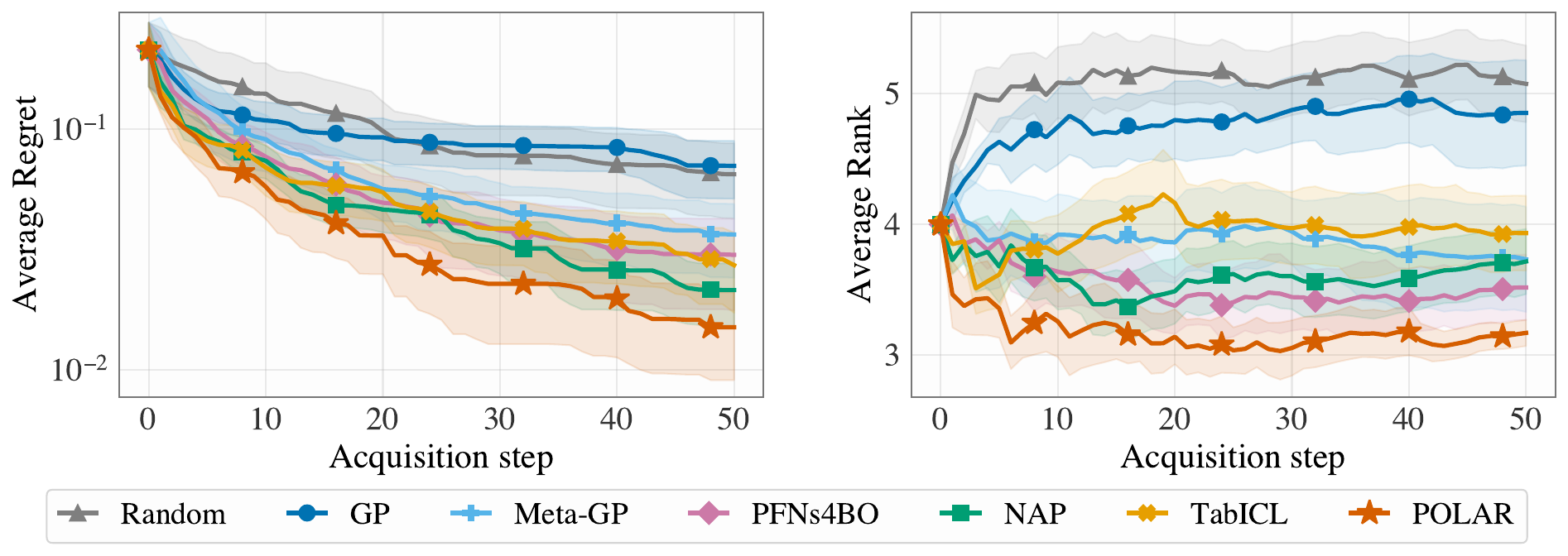}
    \vspace{-2mm}
    \caption{\textbf{Hyperparameter optimisation on HPO-B.} Average regret  (left) and average rank (right) aggregated across the six search spaces. Shaded regions denote one standard error across the test tasks. \method achieves the lowest regret and the best average rank throughout the acquisition trajectory.}
    \label{fig:hpo}
    \vspace{-2mm}
\end{figure}

\subsection{Molecular docking optimisation
} \label{sec:dockstring}

\begin{wrapfigure}{r}{0.37\textwidth}
\vspace{-7mm}
    \begin{center}
        \includegraphics[width=\linewidth]{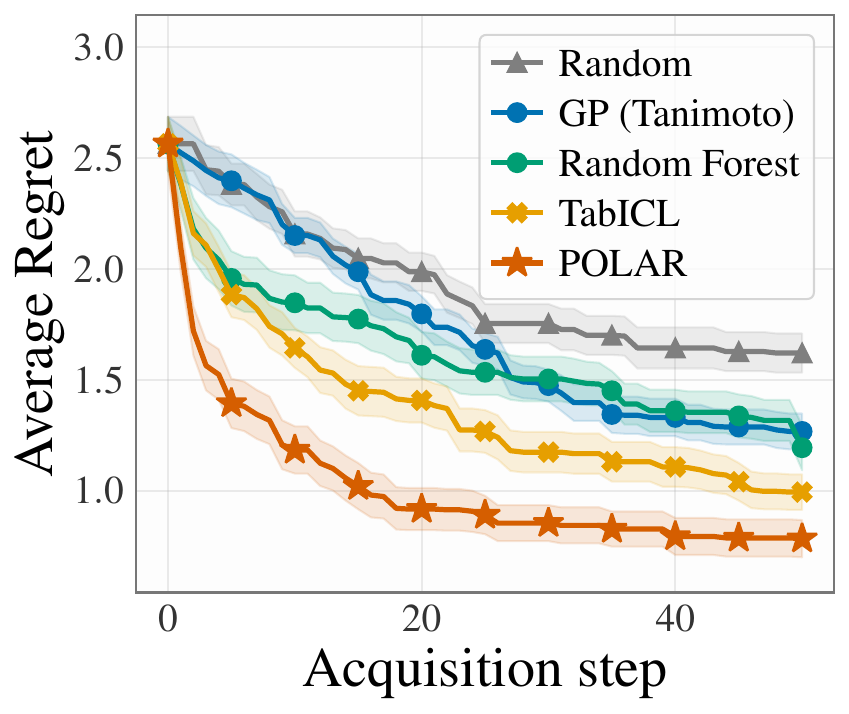}
    \end{center}
    \vspace{-4mm}
    \caption{\textbf{\textsc{Dockstring}}. Average regret (mean $\pm$ s.e.) across the six held-out targets.} 
    \label{fig:dockstring}
\vspace{-6mm}
\end{wrapfigure}

Finally, we evaluate our method on a real-world high-dimensional molecular optimisation benchmark, \textsc{Dockstring} \citep{garcia2022dockstring}, which provides docking scores for over 260,000 molecules against a panel of protein targets. 
Since the physicochemical interactions that govern binding affinity are inherently correlated across these structurally related targets, this dataset serves as a natural testbed for meta BO.

We formulate the task as identifying the molecule that minimises the docking score for a given protein target. We train our policy on a meta-training set of 8 protein targets and evaluate its zero-shot transfer performance on a held-out test set of 6 distinct targets; full target lists are provided in Appendix~\ref{app:dockstring}. Each molecule is represented by a 512-bit Morgan fingerprint computed from its SMILES string.

Operating in a 512-dimensional space presents significant challenges for traditional BO. Standard GPs equipped with continuous kernels (e.g., RBF) notoriously struggle in high-dimensional, discrete spaces. Instead, we equip the GP with a Tanimoto kernel, which measures set overlap between binary fingerprints and is the standard choice for fingerprint-based molecular BO \citep{griffiths2023gauche}. 
We additionally compare against a Random Forest surrogate, a standard baseline in cheminformatics that is commonly paired with molecular fingerprints \citep{svetnik2003random, yang2019analyzing}. For the amortised surrogate baseline, we use exactly the same finetuned TabICL backbone as in our method.

\paragraph{Results.} \Cref{fig:dockstring} reports average regret across the six test targets. \method achieves the lowest regret throughout the trajectory. Both \method and TabICL substantially outperform the non-amortised GP and Random Forest baselines, indicating that meta-training across targets allows the backbone to capture shared structure. The further gap between \method and TabICL confirms that learning the acquisition rule directly from the belief representation outperforms a hand-designed acquisition.

\section{Conclusions} \label{sec:discussion}

\looseness=-1
We introduced \method, an amortised framework for adaptive data acquisition based on a decision-theoretic view: policies need not act directly on raw histories or explicit posterior approximations, but can instead act on belief representations. By instantiating these representations with pretrained predictive foundation models, \method decouples belief representation from decision-making and trains an acquisition policy on them. Across tasks, \method surpasses classical non-amortised methods and state-of-the-art amortised baselines while requiring up to $100\times$ fewer training samples.

\begin{ack}
DH, LA, and SK were supported by the Research Council of Finland (Flagship programme: Finnish Center for Artificial Intelligence FCAI, 359207). CH was supported by Business Finland (VirtualLab4Pharma, grant agreement 3597/31/2023) and the European Union (Horizon Europe, grant agreement 101214398, ELLIOT). 
LA was also supported by Research Council of Finland grants 358980 and 356498. SK was also supported by the UKRI Turing AI World-Leading Researcher Fellowship, [EP/W002973/1]. ZH is supported by the EPSRC CDT in Statistics and Machine Learning (EP/Y034813/1). TR is supported by the EPSRC grant EP/Y037200/1.

We acknowledge CSC – IT Center for Science, Finland, for computational resources provided by the LUMI supercomputer, owned by the EuroHPC Joint Undertaking and hosted by CSC and the LUMI consortium (LUMI projects 462000943 and 462000874). Access was provided through the Finnish LUMI-OKM allocation. We also acknowledge the computational resources provided by the Aalto Science-IT Project from Computer Science IT.
\end{ack}

\bibliography{bibliography}
\bibliographystyle{apalike}

\newpage
\appendix

\setcounter{figure}{0}
\setcounter{table}{0}
\setcounter{equation}{0}
\setcounter{algorithm}{0}
\renewcommand{\thefigure}{A\arabic{figure}}
\renewcommand{\theequation}{A\arabic{equation}}
\renewcommand{\thetable}{A\arabic{table}}
\renewcommand{\thealgorithm}{A\arabic{algorithm}}
\renewcommand{\theHfigure}{A\arabic{figure}}
\renewcommand{\theHtable}{A\arabic{table}}
\renewcommand{\theHalgorithm}{A\arabic{algorithm}}

{
\begin{center}
\Large
    \textbf{Appendix}
\end{center}
}

The appendix is organized as follows: 
\begin{itemize}
    \item In Appendix~\ref{app:method}, we provide additional details of \method, including the backbone architectures, the policy head, and the training details.
    \item In Appendix~\ref{app:exp_details}, we describe the experimental setups in detail, covering task specifications, baseline implementations, and evaluation protocols for the Bayesian experimental design, hyperparameter optimisation, and molecular docking benchmarks.
    \item In Appendix~\ref{app:al_exp}, we extend \method to a loss-driven active learning task.
    \item In Appendix~\ref{app:additional_exps}, we report additional benchmark results and ablations examining belief representations, task-score estimation, temporal credit assignment, and transfer under controlled distribution shifts.
    \item In Appendix~\ref{app:resource_and_license}, we provide an overview of the computational resources and software dependencies used in this work.
\end{itemize}

\section{Additional details about \method} \label{app:method}

\subsection{Backbone}

The default backbone in our experiments is the TabICLv2 regression variant \citep{qu2026tabiclv2}. TabICLv2 is a transformer-based in-context learner pretrained entirely on synthetic tabular data. Architecturally, TabICLv2 inherits the three-stage compress-then-ICL design of TabICLv1 \citep{qu2025tabicl}: a column-wise transformer that embeds each feature in isolation, a row-wise transformer that aggregates feature embeddings into a single fixed-dimensional representation per row, and a dataset-wise transformer that performs in-context learning across the entire context-query set. TabICLv2 extends this design with target-aware row embeddings, repeated feature grouping, and a query-aware scalable softmax that improves generalisation to longer contexts than those seen during pretraining; the model is pretrained with the Muon optimiser on a large corpus of synthetic datasets generated to maximise prior diversity. We use the official regression checkpoint released by the authors, which is trained to predict 999 conditional quantiles of $y^\star$ via the aggregated pinball loss. As a robustness check, we additionally evaluate \method with the TabPFNv2.5 backbone \citep{grinsztajn2025tabpfn}, using the public \texttt{tabpfn-v2.5-regressor-v2.5\_default.ckpt} checkpoint; results are reported in Appendix~\ref{app:location_finding_exp}.

The auxiliary prediction loss aligns the backbone with the current task distribution by supervising its predictive head on target points sampled within each trajectory. The exact form depends on which quantity the backbone is trained to predict.

TabPFNv2.5 outputs a categorical distribution over discretised output bins, which we treat as an approximate predictive density. The auxiliary loss is the standard negative log-likelihood
\begin{equation}
\ell_{\mathrm{pred}}^{\mathrm{NLL}} = - \log q_\psi(y_m \mid x_m, \mathcal{D}_{t-1}).
\end{equation}

TabICLv2 outputs $K = 999$ conditional quantile estimates $\hat{q}_\psi^{(\tau_k)}(x_m, \mathcal{D}_{t-1})$ for evenly spaced quantile levels $\tau_k = k / (K+1)$, $k=1, \ldots, K$. We follow the original training objective and adopt the aggregated pinball loss
\begin{equation}
\ell_{\mathrm{pred}}^{\mathrm{pinball}}\bigl(q_\psi(\cdot \mid x_m, \mathcal{D}_{t-1}), \, y_m\bigr) = \frac{1}{K} \sum_{k=1}^{K} \rho_{\tau_k}\!\bigl(y_m - \hat{q}_\psi^{(\tau_k)}(x_m, \mathcal{D}_{t-1})\bigr),
\end{equation}
where $\rho_\tau(u) = \max(\tau u, (\tau - 1) u)$ is the standard pinball (tilted absolute) loss for quantile level $\tau$. In all experiments, the policy loss in \Cref{eq:policy_loss} and the auxiliary prediction loss are weighted equally, $\lambda_{\mathrm{pol}} = \lambda_{\mathrm{pred}} = 1$. 
For density-based rewards and NLPD evaluation in Appendix~\ref{app:al_exp}, following \citet{qu2026tabiclv2}, we convert the predicted quantiles into a continuous distribution by enforcing monotonicity, linearly interpolating the quantile function between adjacent levels, extrapolating exponential tails beyond the extreme predicted quantiles, and then evaluating \(\log q_\psi(y\mid x, \mathcal{D}) = -\log \partial_\tau Q_\psi(\tau\mid x, \mathcal{D})\vert_{\tau = F_\psi(y\mid x, \mathcal{D})}\); the predictive mean is computed as \(\mu_\psi(x, \mathcal{D})=\int_0^1 Q_\psi(\tau\mid x, \mathcal{D})\,d\tau\).

\subsection{Policy head}

The default policy head used throughout the main experiments is a lightweight MLP with two hidden layers, hidden width 128, and GELU activations. It maps each candidate-conditioned representation, concatenated with the normalised remaining budget, to a scalar logit, and the policy distribution over candidates follows the softmax.

For the loss-driven active learning experiments in Appendix~\ref{app:al_exp}, where the goal is to optimise predictive performance over a user-specified target set, we use a target-aware variant of the policy head. This variant uses a single transformer encoder layer (4 attention heads, hidden width 1024, GELU activation), allowing each candidate token to attend to the target-point representations produced by the backbone. 

During training, we optimise the policy head and the backbone using AdamW. Gradients are clipped to unit global norm. We sample candidates stochastically from the policy distribution to ensure sufficient exploration during training. At test time, we instead select the argmax candidate. 

\section{Experimental details} \label{app:exp_details}

\subsection{Benchmarking on Bayesian experimental design tasks} \label{app:bed}

\textbf{Location finding} \citep{sheng2005maximum} is a widely adopted benchmark for sequential BED \citep{foster2019variational, foster2021deep, ivanova2021implicit, huang2025aline}. The objective is to recover the positions of $K$ unknown sources in $\mathbb{R}^d$, denoted $\theta = \{\theta_k \in \mathbb{R}^d\}_{k=1}^K$, by adaptively choosing measurement locations $x \in \mathbb{R}^d$ at which noisy signal intensities are recorded. Each source radiates a signal whose magnitude attenuates with distance under an inverse-square law, so that a measurement taken at location $x$ aggregates contributions from every source:
\begin{equation}
\mu(\theta, x) \;=\; b \;+\; \sum_{k=1}^K \frac{\alpha_k}{m + \lVert \theta_k - x \rVert^2},
\end{equation}
where the constants $\alpha_k$ specify the strength of each source, while $b > 0$ and $m > 0$ govern the background offset and the saturation behaviour of the signal at short range, respectively. The experimenter does not observe $\mu$ directly; instead, the log-intensity is observed under additive Gaussian noise:
\begin{equation}
\log y \mid \theta, x \;\sim\; \mathcal{N}\!\big(\log \mu(\theta, x), \sigma^2\big).
\end{equation}
Following standard practice, we instantiate the task with $K=2$, $\alpha_k=1$, $b=0.1$, $m=10^{-4}$, and $\sigma=0.5$. Designs are constrained to the unit hypercube $x \in [0,1]^d$, and the prior over each source places independent uniform distributions on every coordinate, $\theta_{k,j} \sim \mathrm{Unif}[0,1]$ for $j=1,\dots,d$.

\textbf{Constant elasticity of substitution} (CES; \citealp[]{arrow1961capital}) originates in economic theory and models how a consumer values bundles of goods. In the experimental design formulation, the experimenter aims to identify a participant's latent preference structure by repeatedly presenting them with pairs of baskets and recording a (potentially noisy) judgement about their relative desirability. A basket $c \in [0,100]^K$ encodes the quantities held of each of $K$ goods. The participant's preferences are captured by latent parameters $\theta=(\rho,\boldsymbol{\alpha},u)$, where $\rho \in (0,1)$ determines the degree of substitutability between goods, $\boldsymbol{\alpha} \in \Delta^{K-1}$ assigns simplex-valued weights to the goods, and $u>0$ modulates the overall responsiveness of the participant's reports. A single experimental design takes the form of a basket pair $x=(c,c') \in [0,100]^{2K}$, and elicits a bounded preference rating $y \in [0,1]$.

The utility assigned to a basket follows the canonical CES form:
$U(c) \;=\; \left(\sum_{i=1}^{K} c_i^{\rho}\,\alpha_i\right)^{\frac{1}{\rho}}$.
We adopt the following prior specification over the latent parameters:
\begin{equation}
\rho \sim \mathrm{Beta}(1,1),\qquad
\boldsymbol{\alpha} \sim \mathrm{Dirichlet}(\mathbf{1}_K),\qquad
\log u \sim \mathcal N(1,3^2).
\end{equation}
For a given query $x=(c,c')$ and parameter setting $\theta$, the underlying (unobserved) utility differential is generated from
\begin{equation}
\eta \sim \mathcal N\!\Big(u\,(U(c)-U(c')),\; u^2\,\tau^2\,(1+\lVert c-c'\rVert)^2\Big),
\end{equation}
and then transformed through a sigmoid link and clipped to a bounded interval to yield the reported outcome:
$y \;=\; \mathrm{clip}\big(\sigma(\eta),\;\epsilon,\;1-\epsilon\big)$,
with $\sigma(\cdot)$ the standard sigmoid. We use $K=3$, $\tau=0.005$, $\epsilon=2^{-22}$, and query budget $T=10$ in all experiments. Since TFMs are pretrained on standardised inputs and outputs, we standardise the features before feeding them to the backbone.

\subsubsection{Baselines} 

\textbf{Deep Adaptive Design} \citep{foster2021deep} learns an amortised design policy by directly maximising the sPCE lower bound on the total EIG over $T$-step trajectories. The policy network consists of a two-layer MLP encoder with hidden width 128 that maps each design-observation pair $(x_i, y_i)$ to a 16-dimensional embedding, a permutation-invariant pooling operation that aggregates these embeddings into a history representation, and a one-layer MLP emitter that maps the pooled representation to the next design. We train with Adam (learning rate $5\times 10^{-5}$, $\beta=(0.8, 0.998)$), gradient clipping at 1.0, and an exponential learning-rate decay with factor 0.98 every 1000 epochs. For location finding, we use $L=2 \times 10^4$ contrastive samples for sPCE estimation during training. For CES, we use $L=10^5$.

\textbf{RL-BOED} \citep{blau2022optimizing} formulates sequential BED as a hidden-parameter Markov decision process and learns an amortised design policy using reinforcement learning. The reward is defined stepwise as the marginal contribution of the newly acquired design-observation pair to the sPCE. We use Randomized Ensembled Double Q-learning (REDQ) to train the continuous-design policy. The history encoder follows a DAD-style permutation-invariant architecture: each concatenated design-observation pair $(x_i,y_i)$ is passed through a two-layer MLP with 128 hidden units and ReLU activations, followed by a 64-dimensional linear output layer; the resulting embeddings are summed to form the history representation. The policy emitter outputs the mean and log-variance of independent Tanh-Gaussian distributions over the design dimensions. The critic networks use the same history encoder and concatenate the encoded history with the candidate design before passing it through a two-layer 128-unit ReLU MLP. We train with $L=10^5$ contrastive samples for the sPCE reward. The remaining RL hyperparameters are listed in \Cref{tab:rlboed_hparams}.

\begin{table}[h]
\centering
\caption{Hyperparameters used for RL-BOED.}
\label{tab:rlboed_hparams}
\begin{tabular}{lcc}
\toprule
Parameter & Location Finding & CES \\
\midrule
Critics $N$ & 2 & 2 \\
Random target subset $M$ & 2 & 2 \\
Discount factor $\gamma$ & 0.9 & 0.9 \\
Target update rate $\tau$ & $10^{-3}$ & $5\times 10^{-3}$ \\
Policy learning rate & $10^{-4}$ & $3\times 10^{-4}$ \\
Critic learning rate & $3\times 10^{-4}$ & $3\times 10^{-4}$ \\
Replay buffer size & $10^7$ & $10^6$ \\
Minimum buffer size & $10^5$ & $10^5$ \\
Entropy-temperature learning rate & $3\times 10^{-4}$ & $3\times 10^{-4}$ \\
\bottomrule
\end{tabular}
\end{table}

\textbf{ALINE} \citep{huang2025aline} jointly amortises sequential design and posterior inference with a shared masked transformer architecture. Each design $x$ and observation $y$ is embedded by separate two-layer MLPs with hidden width 128 into 32-dimensional representations; context tokens are formed by summing the design and observation embeddings, candidate-query designs are represented by design embeddings alone, and one learnable target token is introduced for each latent parameter dimension. These tokens are processed by a 3-layer transformer encoder with model width 32, feedforward width 128, and 4 attention heads. A one-hidden-layer MLP acquisition head maps the query representations to a categorical distribution over the candidate design pool, while a separate one-hidden-layer MLP posterior head predicts a 10-component Gaussian mixture for each target parameter token. We train ALINE with a joint objective consisting of a negative log-likelihood term for posterior prediction and a REINFORCE-style design loss based on one-step improvements in target log-likelihood, weighted equally with $\alpha=1$ and discount factor $\gamma=1$. For the sample-efficiency experiments, we use AdamW with cosine annealing and layer-wise learning rates of $10^{-3}$ for the acquisition and posterior heads and $2\times 10^{-4}$ for the shared embedder and transformer.

\subsubsection{Evaluation details}

Following \citet{foster2021deep}, we evaluate all amortised policies using the sequential Prior Contrastive Estimation (sPCE) lower bound and the sequential Nested Monte Carlo (sNMC) upper bound on the total expected information gain. Given a sampled trajectory $(\theta_0, h_T) \sim p(\theta, h_T \mid \pi)$ together with $L$ contrastive samples $\theta_{1:L} \sim p(\theta)$ drawn independently from the prior, the sPCE and sNMC estimators are defined as
\begin{equation}
\mathcal{L}_T(\pi, L) = \mathbb{E}\!\left[\log \frac{p(h_T \mid \theta_0, \pi)}{\frac{1}{L+1}\sum_{\ell=0}^{L} p(h_T \mid \theta_\ell, \pi)}\right], \quad
\mathcal{U}_T(\pi, L) = \mathbb{E}\!\left[\log \frac{p(h_T \mid \theta_0, \pi)}{\frac{1}{L}\sum_{\ell=1}^{L} p(h_T \mid \theta_\ell, \pi)}\right],
\end{equation}
where the outer expectation is taken over $\theta_0, h_T \sim p(\theta, h_T \mid \pi)$ and $\theta_{1:L} \sim p(\theta)$. Both bounds become tight as $L \to \infty$ at a rate $O(L^{-1})$ \citep{foster2021deep}. For training, \method uses $L=10^5$ contrastive samples for both tasks. For the evaluations, we use $L=10^6$ contrastive samples for location finding and $L = 10^7$ for CES.

We report performance against the total number of training samples consumed during policy learning, where a "sample" denotes a single design-observation pair $(x, y)$ produced by the simulator. This convention provides a uniform unit of comparison that is independent of architectural choices such as batch size, replay-buffer capacity, or auxiliary loss terms.

For DAD and ALINE, every gradient step consumes one fresh batch of trajectories of length $T$, so the cumulative number of training samples after $E$ epochs is $E \cdot B \cdot T$, where $B$ is the batch size and $T$ is the experiment horizon. For RL-BOED, which maintains a replay buffer, only the trajectories newly synthesised at each epoch contribute to the sample count; the batch size of the RL update governs how many transitions are drawn from the buffer per gradient step, but does not generate new simulator data, and is therefore excluded from the accounting. Concretely, if RL-BOED rolls out $B_{\mathrm{env}}$ new trajectories per epoch, the cumulative count after $E$ epochs is $E \cdot B_{\mathrm{env}} \cdot T$.

For \method, the auxiliary prediction loss requires evaluating the backbone on $M$ target points per trajectory in addition to the $T$ design-observation pairs visited by the policy. Each target point corresponds to a fresh sample drawn from the simulator and therefore contributes to the total sample budget. We thus account for both sources, giving a per-epoch consumption of $B \cdot (T + M)$ and a cumulative count of $E \cdot B \cdot (T + M)$ after $E$ epochs. We use $M=20$ target points and $B=50$ trajectories per update for location finding, and $M=10$ and $B=20$ for CES. Our sample-efficiency comparison measures target-task simulator samples consumed during policy learning, conditional on access to an existing pretrained backbone; it does not include foundation-model pretraining. We consider this accounting appropriate because the backbones are pretrained entirely on synthetic tabular data that is task-agnostic, cheap to generate, and shared across all downstream applications. Once pretrained, the same checkpoint is reused without modification across BED, BO, and AL tasks, so the pretraining cost is amortised across all downstream uses rather than attributable to any single task. The reported gains should be interpreted as improvements in target-task sample efficiency, rather than reductions in total data or compute over the full pretraining-and-adaptation lifecycle.

Both \method and ALINE operate over a finite candidate pool at each acquisition step, whereas DAD and RL-BOED produce continuous designs directly. To match the original formulations of all baselines while keeping the pool-based methods well-resourced, we use a candidate pool of $|\mathcal{C}| = 2000$ designs sampled uniformly from the design space at the start of each trajectory for location finding, and $|\mathcal{C}| = 20000$ for CES. During training, \method uses smaller pools of $|\mathcal{C}|=1000$ and $|\mathcal{C}|=2000$, respectively.

\subsection{Hyperparameter optimisation} \label{app:hpo}

\subsubsection{Task description}

HPO-B \citep{arango2hpo} is a large-scale meta-dataset for black-box hyperparameter optimisation (HPO) assembled from the OpenML repository. It contains 6.4 million hyperparameter evaluations. We use the curated HPO-B-v3 split, which retains the 16 search spaces and provides predefined meta-train, meta-validation, and meta-test partitions, together with five fixed initialisation seeds per test task to support reproducible comparisons. Each search space corresponds to the hyperparameter space of a particular machine-learning algorithm (e.g., SVM, XGBoost, glmnet), and each meta-dataset within a search space corresponds to evaluations on a specific tabular dataset. The hyperparameter ranges are normalised to $[0,1]^d$ and categorical hyperparameters are one-hot encoded. Following \citet{maraval2023end}, we evaluate on six search spaces that together cover a representative range of underlying model families and input dimensionalities: \texttt{glmnet} (search space 5860, $d=2$), \texttt{rpart.preproc} (search space 4796, $d=3$), \texttt{rpart} (search space 5859, $d=6$), \texttt{ranger} (search space 5889, $d=6$), \texttt{svm} (search space 5527, $d=8$), and \texttt{xgboost} (search space 5906, $d=16$). We train on the provided meta-training datasets and evaluate on the held-out meta-test tasks. We adopt the recommended evaluation protocol of \citet{arango2hpo}: each test task is run with the five fixed initialisation seeds for $T=50$ acquisition steps, and we report the average normalised regret and average rank as defined in \citet{arango2hpo}.

\subsubsection{Baselines} 

\paragraph{GP.} A standard Gaussian process surrogate fitted directly on the test task, with no use of meta-training data. We use an exact GP with a constant mean, a scaled RBF kernel, and a Gaussian likelihood. The kernel and noise hyperparameters are optimised by maximizing the exact marginal likelihood on the observed configurations after each BO round.

\paragraph{Meta-GP.} Following \citet{maraval2023end}, a meta-trained Gaussian process pretrains the RBF kernel hyperparameters on the meta-training datasets of the corresponding search space and uses these pretrained values to initialise the GP at test time. Subsequent acquisitions then refit the kernel hyperparameters on the growing test-time observation set, starting from this meta-learned initialisation rather than from a generic prior.

\paragraph{PFNs4BO \citep{muller2023pfns4bo}.} PFNs4BO uses Prior-data Fitted Networks \citep{muller2021transformers} as transformer-based surrogates for BO: a single transformer is meta-trained offline to approximate the posterior predictive distribution of a chosen prior over functions, after which inference for any new context set is performed in a single forward pass without further fitting. The strongest variant of the model is trained on the so-called HEBO+ prior, which augments the HEBO Gaussian process prior \citep{cowen2022hebo} with input and output warping and additional hyperparameter randomisation, yielding a surrogate that has been shown to outperform standard GP surrogates on the HPO-B benchmark. We use the publicly released \texttt{hebo-plus-model} checkpoint from the official PFNs4BO repository without any further finetuning on the meta-training tasks.

\paragraph{NAP \citep{maraval2023end}.} NAP is an end-to-end differentiable meta-BO framework that jointly meta-learns a transformer-based neural process \citep{nguyen2022transformer} surrogate and an acquisition function via reinforcement learning, with an auxiliary supervised likelihood loss that stabilises training and provides an inductive bias toward valid probabilistic predictions. NAP is the closest amortised baseline to \method, in that both methods learn the acquisition rule rather than relying on a hand-designed one. Since the HPO-B search spaces, test tasks, and initialisation seeds used by \citet{maraval2023end} are identical to ours, we directly use the per-step regret values reported in their official release for fair comparison.

\paragraph{TabICL \citep{qu2026tabiclv2}.} A surrogate-based ablation that uses the same finetuned TabICL backbone as \method, but discards the learned policy head and instead pairs the backbone's predictive distribution with the EI acquisition function. This baseline isolates the contribution of learning the acquisition rule from the contribution of the meta-learned belief representation: the difference between TabICL and \method can be attributed entirely to replacing a hand-designed acquisition with one trained on top of the same belief representation.

\subsubsection{Evaluation details}

For each HPO-B test task, all methods start from the same official initialisation and are evaluated on the same candidate pool. We report the normalised regret
\begin{equation}
    r_t =
    \frac{f^\star - f_t^{\mathrm{best}}}{f^\star - f_{\min}},
\end{equation}
where $f_t^{\mathrm{best}}$ is the best objective value observed by the method up to step $t$, $f^\star$ is the best value in the full task pool, and $f_{\min}$ is the worst value in the same pool. We average normalised regret over test datasets and initialisation seeds.

We also report the normalised rank to compare methods across tasks with different regret scales. For each task, seed, and optimisation step, methods are ranked by their normalised regret, with lower regret receiving a better rank and ties assigned the average rank. These ranks are then averaged over datasets, seeds, and search spaces.

\subsection{Molecular docking optimisation} \label{app:dockstring}

\subsubsection{Task description}

\textsc{Dockstring} \citep{garcia2022dockstring} is a benchmark for molecular optimisation built around AutoDock Vina \citep{trott2010autodock} docking simulations against a curated panel of medically relevant protein targets. The full dataset comprises docking scores and binding poses for over 260000 drug-like molecules evaluated against 58 targets, where each docking score quantifies the predicted binding affinity between a ligand and a target. Lower scores indicate stronger predicted binding, so the optimisation objective is to identify molecules that minimise the docking score for a given target. We frame this as a pool-based BO problem. Each molecule is represented by its SMILES string and converted to a 512-bit Morgan fingerprint with radius 2. We randomly subsample 10000 molecules from the full dataset and use this fixed subset throughout all experiments. We split the 14 protein targets used in our experiments into a meta-training set of 8 targets, JAK2, KIT, LCK, SRC, IGF1R, ABL1, MET, EGFR, and a held-out test set of 6 targets, PTK2, FGFR1, CSF1R, CDK2, MAPK14, KDR. All 14 are protein kinases, which share a conserved ATP-binding pocket and substantial structural homology across the family. This shared geometry means that the physicochemical features predictive of strong binding are correlated across targets, making the kinase family a natural setting for meta-learning.

\subsubsection{Baselines} 

\paragraph{GP with Tanimoto kernel.} Standard continuous kernels such as the RBF or Matérn kernels are poorly suited to high-dimensional binary fingerprint inputs, where Euclidean distance is a weak measure of chemical similarity. We instead equip the GP with the Tanimoto kernel, a standard choice for fingerprint-based molecular BO \citep{griffiths2023gauche}. For two binary fingerprints $x, x' \in \{0,1\}^d$, the Tanimoto kernel is defined as
\begin{equation}
k_{\mathrm{Tanimoto}}(x, x') = \sigma^2 \cdot \frac{\langle x, x' \rangle}{\|x\|^2 + \|x'\|^2 - \langle x, x' \rangle},
\end{equation}
where $\sigma^2$ is a learnable output scale and $\langle \cdot, \cdot \rangle$ denotes the inner product. The kernel returns 1 for identical fingerprints and 0 for fingerprints with no shared bits, providing a meaningful similarity signal for sparse binary representations. We refit the kernel hyperparameters by maximum marginal likelihood after each acquisition step.

\paragraph{Random Forest.} Random forests are a standard surrogate in cheminformatics and are commonly paired with molecular fingerprints \citep{svetnik2003random, yang2019analyzing}. We use an ensemble of 256 regression trees with bootstrapped training sets and the standard $\sqrt{d}$ feature-subsampling rule at each split. The surrogate is refit on the full set of observed molecules at each acquisition step.

\paragraph{TabICL.} We use the same setup as we used in our HPO-B experiments, where the same finetuned TabICL backbone as \method is used but replaces the learned policy head with a hand-designed acquisition function.

\subsubsection{Evaluation protocol}

For each of the 6 held-out test targets, we evaluate all methods under the same protocol: 10 randomly sampled molecules form the initial context, after which the policy selects $T = 50$ acquisitions, with a candidate pool of size $|\mathcal{C}| = 2000$. We report the average regret and aggregate by averaging across the 6 test targets and 5 random seeds for each test target.

\section{Loss-driven active learning} \label{app:al_exp}

In this section, we evaluate \method on a loss-driven active learning task, where the goal is to select query points that improve predictive performance on a downstream task distribution $p^\star$ under a user-specified loss.

Active learning seeks to acquire labelled data that maximises the predictive performance of a downstream model. In practice, a clinician may care primarily about predicting high-risk patients accurately, a structural engineer about responses near a failure threshold, and a drug discovery pipeline about molecules with high binding affinity. Standard information-theoretic acquisition objectives such as BALD \citep{houlsby2011bayesian} and EPIG \citep{smith2023prediction} treat all outcomes symmetrically and therefore cannot directly target such loss-weighted criteria. \citet{huang2026loss} formalise this gap and derive a principled extension via weighted Bregman divergences, yielding myopic acquisition rules such as weighted variance reduction (GP-VR$_w$) that explicitly target a user-specified weighted loss. We adopt the same loss-driven framing but instantiate it in our amortised setting.

Concretely, we consider pool-based active learning under a weighted loss $\ell_\omega(y, q) =\omega(y) \ell(y,q)$, where $\omega: \mathcal{Y} \to \mathbb{R}_+$ encodes which outcomes matter most for the downstream task. At round $t$, the learner observes a context $\mathcal{D}_{t-1} = \{(x_i, y_i)\}_{i=1}^{t-1}$, selects a query $x_t$ from a finite pool $\mathcal{C}_t$, and observes $y_t \sim p(y \mid x_t, \theta)$ where $\theta$ is the latent function. After $T$ rounds, performance is evaluated on a held-out target set drawn from the target input distribution $p_\star(x^\star)$. 

\method instantiates this setting using the same two-component architecture as in our other experiments. TFM plays the role of an amortised surrogate model: given the current context $\mathcal{D}_{t-1}$ and a query input $x^\star$, a single forward pass yields the approximate posterior predictive $q_\psi(y^\star \mid x^\star, \mathcal{D}_{t-1})$, replacing the GP surrogate that classical AL methods refit at every round. The policy head then scores each candidate in $\mathcal{C}_t$ from the same forward pass and samples the next query $x_t$.
For the utility, we use the weighted one-step improvement in predictive log-density on the target set,
\begin{equation}
\label{eq:al-reward}
u_t = \mathbb{E}_{x^\star \sim p^\star}\big[\omega(y^\star)\big(\log q_\psi(y^\star \mid x^\star, \mathcal{D}_t) - \log q_\psi(y^\star \mid x^\star, \mathcal{D}_{t-1})\big)\big],
\end{equation}
which measures the weighted improvement in the backbone's own predictive log-density on the target set after observing the new query, and which generalises the dense sEPIG reward of \citet{huang2025aline} from the unweighted to the weighted setting. In our experiments, we adopt the exponential weighting $\omega(y) = \exp(\beta y)$ with $\beta = 10$, prioritising regions of high outcome value. At deployment, no GP fitting or acquisition optimisation is required: the next query is produced in a single forward pass per round.

\subsection{Training distribution}

We train \method on synthetic functions drawn from a distribution over GPs with randomised kernels and hyperparameters, following a procedure similar to ALINE \citep{huang2025aline}. Each task is sampled by:
\begin{enumerate}
    \item Drawing a kernel uniformly from \{RBF, Mat\'ern-3/2, Mat\'ern-5/2\};
    \item Drawing a length-scale $\ell \sim \text{LogUniform}(0.1, 2.0)$ and an output scale $\sigma_f \sim \text{Uniform}(0.1, 1.0)$;
    \item Sampling a function $f \sim \mathcal{GP}(0, k_{\ell, \sigma_f})$.
\end{enumerate}
Before each forward pass, we normalise each function using task-level statistics computed independently of the observed context: for inputs, we use the known domain bounds, and for outputs, we estimate the normalisation mean and variance from a large set of reference points sampled from that function. The horizon is $T = 20$, with an initial context of 2 points and a candidate pool of size $K = 200$.

\subsection{Baselines}

We compare against four non-amortised GP baselines.

\textbf{GP-RS} (random sampling) selects $x_t$ uniformly at random from the pool.

\textbf{GP-US} (uncertainty sampling) selects the candidate with the highest posterior predictive variance,
\begin{equation}
\text{US}(x) = \mathbb{V}[y \mid x, \mathcal{D}_{t-1}].
\end{equation}

\textbf{GP-VR} (variance reduction) selects the candidate that maximally reduces total predictive variance over the target set $\{x^\star_m\}_{m=1}^{M}$:
\begin{equation}
\text{VR}(x) = \sum_{m=1}^{M} \frac{\text{Cov}_{t-1}(x^\star_m, x)^2}{\mathbb{V}[y \mid x, \mathcal{D}_{t-1}]},
\end{equation}
where $\text{Cov}_{t-1}(x^\star, x)$ is the GP posterior covariance between the latent function values at $x^\star$ and $x$ given $\mathcal{D}_{t-1}$.

\textbf{GP-VR$_\omega$} (weighted variance reduction; \citealp{huang2026loss}) is the loss-matched counterpart of GP-VR, derived from the same weighted-Bregman framework as our reward. It targets the weighted predictive variance under the reweighted predictive $p_\omega(y \mid x^\star) \propto \omega(y) p(y \mid x^\star)$ rather than the unweighted variance.

\subsection{Evaluations} 

We evaluate on two distributions of test functions.

\textbf{GP synthetic.} We sample 100 functions from the same generative procedure used during training. This measures in-distribution performance.

\textbf{Benchmark functions.} We additionally evaluate on four standard regression benchmarks to measure out-of-distribution generalisation: Forrester, Gramacy-Lee, Higdon, and Sine-Gaussian Bumps. The functional forms are:
\begin{align*}
f_{\text{Forr}}(x) &= (6x - 2)^2 \sin(12x - 4), \\
f_{\text{GL}}(x) &= \frac{\sin(10\pi x)}{2x} + (x - 1)^4, \\
f_{\text{Higdon}}(x) &= \sin(2\pi x / 10) + 0.2 \sin(2\pi x / 2.5), \\
f_{\text{SGB}}(x) &= 2\sin(2x) + 8 \phi_{2.5, 0.5}(x) + 10 \phi_{7.5, 0.25}(x) - 6 \phi_{-4.5, 0.5}(x).
\end{align*}
where $\phi_{\mu, \sigma}$ is a Gaussian density with mean $\mu$ and standard deviation $\sigma$. Before evaluation, we apply a fixed per-function normalisation: inputs are centred and scaled using the benchmark domain, and outputs are standardised using statistics estimated from Sobol samples over the same domain. All methods, including GP and \method, are evaluated on these same normalized benchmark versions.

For each benchmark function, we run 100 independent seeds. Each seed draws an initial context of 2 points, a candidate pool of 200 points, and a target set of 100 points; all three sets are sampled uniformly. Performance is reported as weighted RMSE and weighted NLPD on the target set:
\begin{equation}
\text{RMSE}_\omega = \sqrt{\frac{\sum_{m=1}^{M} \omega(y^\star_m)(y^\star_m - \mu(x^\star_m))^2}{\sum_{m=1}^{M} \omega(y^\star_m)}}, \qquad
\text{NLPD}_\omega = -\frac{\sum_{m=1}^{M} \omega(y^\star_m)\log q(y^\star_m \mid x^\star_m)}{\sum_{m=1}^{M} \omega(y^\star_m)},
\end{equation}
where $\mu(x^\star)$ is the predictive mean under the surrogate at the end of the acquisition trajectory.

\subsection{Results} 

\begin{figure}
  \centering
    \includegraphics[width=\linewidth]{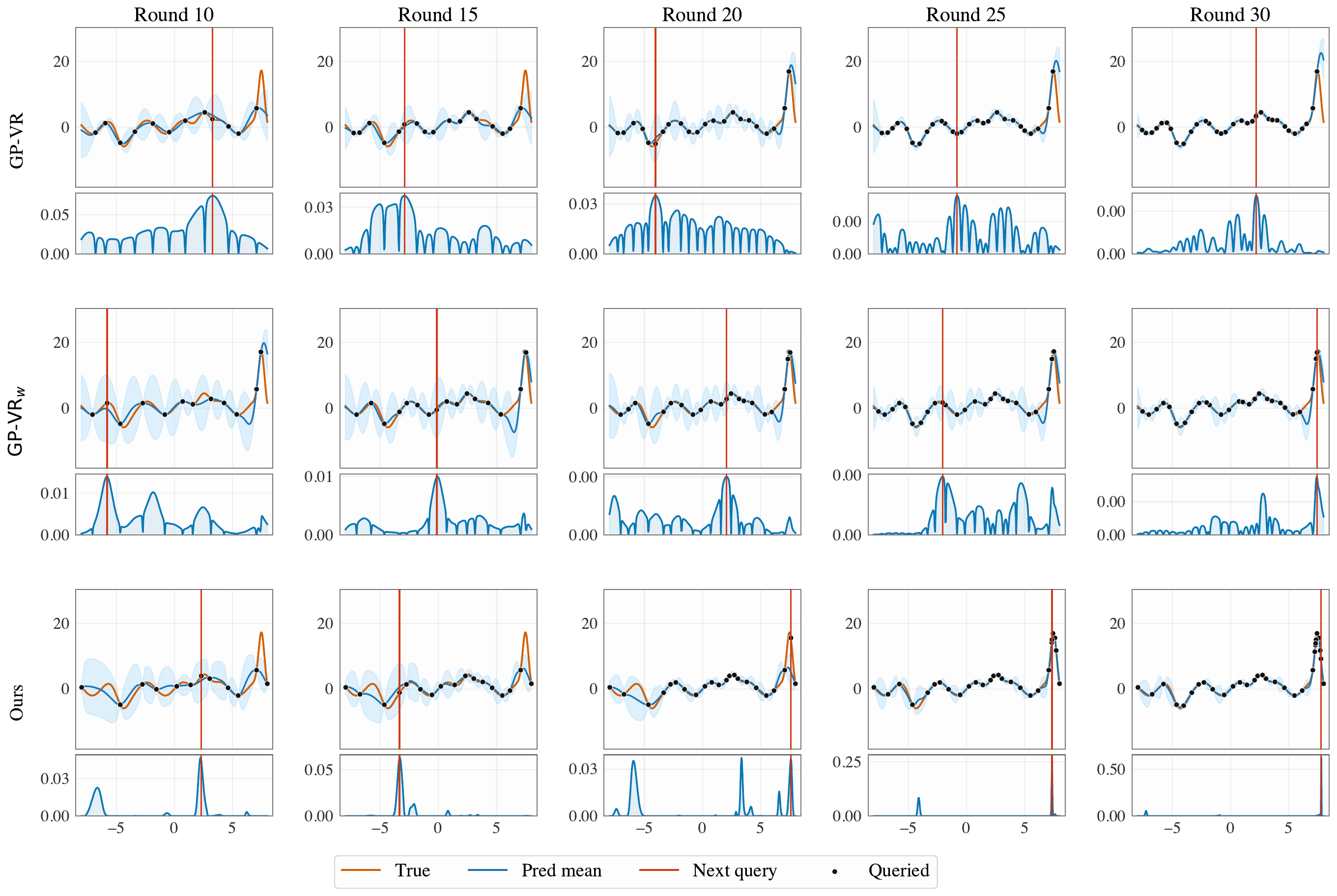}
  \caption{\textbf{Loss-driven active learning}. Acquisition behaviour on the Sine-Gaussian Bumps function from \citet{huang2026loss}. The loss-agnostic GP-VR (top) spreads queries across the domain, while GP-VR$_\omega$ and \method (middle, bottom) concentrate queries near the high-value region.}
  \label{fig:al_vis}
\end{figure}

\Cref{tab:al} reports final-step weighted RMSE and weighted NLPD on both evaluation distributions. First, the two loss-aware methods GP-VR$_\omega$ and \method substantially outperform the loss-agnostic baselines (GP-RS, GP-US, GP-VR) on both metrics and both evaluation distributions. This confirms the central message of \citet{huang2026loss}: when the downstream loss assigns non-uniform weight to outcomes, acquisition rules that ignore the weighting are systematically suboptimal. Second, \method consistently outperforms GP-VR$_\omega$ across both evaluation distributions, suggesting that learning the acquisition rule directly from data outperforms a hand-designed one. \Cref{fig:al_vis} visualises the acquisition behaviour of GP-VR, GP-VR$_\omega$, and \method on the Sine-Gaussian Bumps function from \citet{huang2026loss}.

\begin{table}[h]
\centering
\caption{\textbf{Loss-driven active learning}. Final-step weighted RMSE and weighted NLPD on GP synthetic functions (in-distribution) and benchmark functions (out-of-distribution).}
\label{tab:al}
\begin{tabular}{lcccc}
\toprule
\multirow{2}{*}{Methods} & \multicolumn{2}{c}{GP Synthetic} & \multicolumn{2}{c}{Benchmark Functions} \\
\cmidrule(lr){2-3} \cmidrule(lr){4-5}
 & \makecell{$\text{RMSE}_\omega$ ($\downarrow$)} & \makecell{$\text{NLPD}_\omega$ ($\downarrow$)} & \makecell{$\text{RMSE}_\omega$ ($\downarrow$)} & \makecell{$\text{NLPD}_\omega$ ($\downarrow$)} \\
\midrule
GP-RS & \vpmse{0.40}{0.06} & \vpmse{-0.18}{0.28} & \vpmse{0.74}{0.06} & \vpmse{3.09}{0.78}  \\
GP-US & \vpmse{0.13}{0.02} & \vpmse{-1.26}{0.16} & \vpmse{0.45}{0.05} & \vpmse{4.20}{0.75} \\
GP-VR & \vpmse{0.17}{0.03} & \vpmse{-1.18}{0.16} & \vpmse{0.33}{0.03} & \vpmse{3.30}{0.78} \\
$\text{GP-VR}_\omega$ & \vpmse{0.11}{0.02} & \vpmse{-1.65}{0.15} & \vpmse{0.15}{0.02} & \vpmse{-0.83}{0.33}  \\
\method (ours) & \vpmse{\textbf{0.09}}{0.03} & \vpmse{\mathbf{-2.07}}{0.13} & \vpmse{\mathbf{0.05}}{0.01} & \vpmse{\mathbf{-1.68}}{0.10} \\
\bottomrule
\end{tabular}
\end{table}

\section{Additional experimental results} \label{app:additional_exps}

\subsection{Location finding} \label{app:location_finding_exp}

\begin{figure}[t]
  \centering
    \subfigure[]{\includegraphics[width=0.33\linewidth]{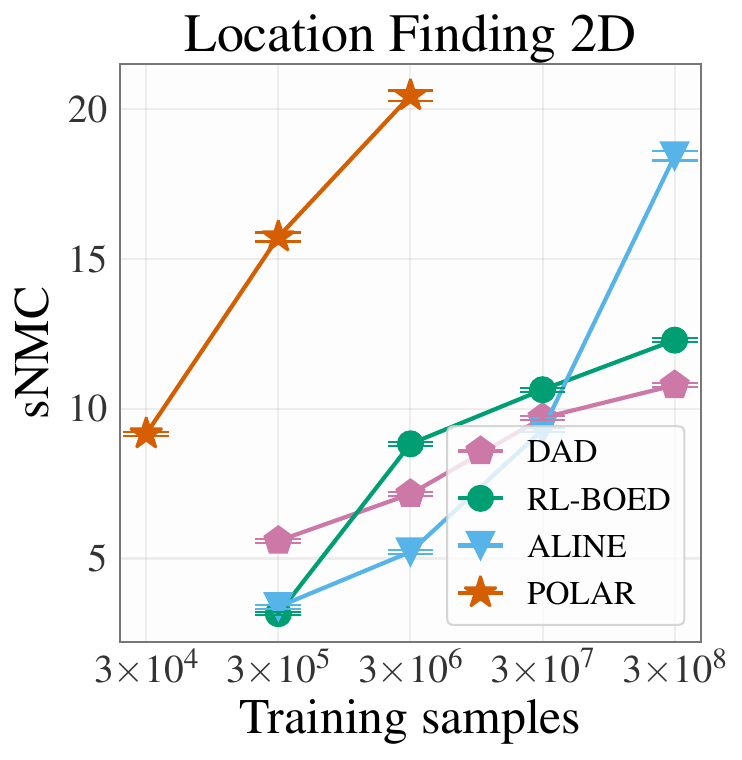}}
    \subfigure[]{\includegraphics[width=0.345\linewidth]{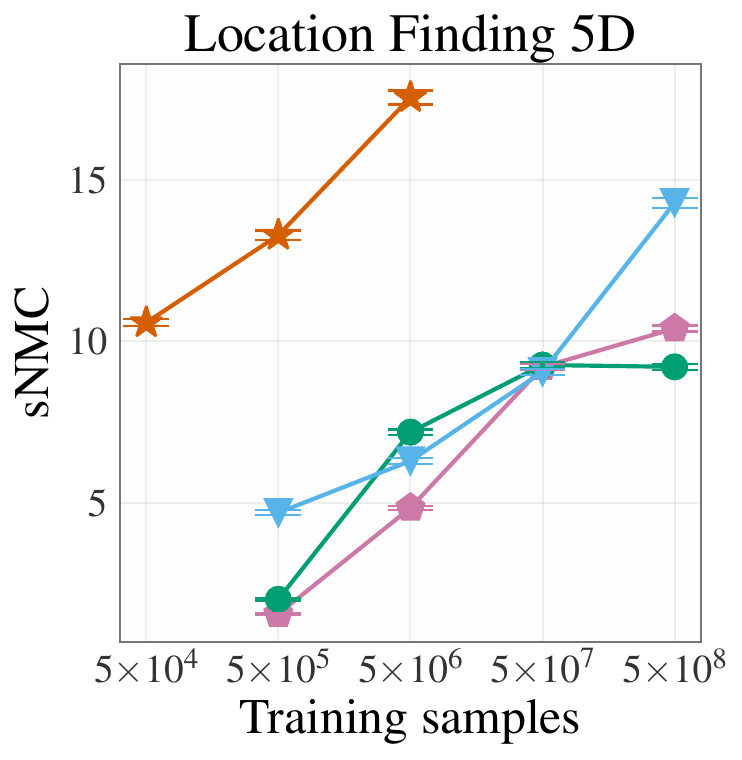}}
  \caption{\textbf{Location finding}. (a) EIG upper bound (sNMC) against total training samples in the 2D setting. Error bars denote standard error across 1000 runs. (b) The same comparison in the 5D setting.}
  \label{fig:location_finding_snmc}
\end{figure}

\textbf{sNMC results.} \Cref{fig:location_finding_snmc}(a) and \Cref{fig:location_finding_snmc}(b) show the sNMC upper bound on EIG against total training samples in the 2D and 5D settings, respectively. The relative ordering of methods mirrors the sPCE comparison in \Cref{fig:location_finding}. 

\begin{figure}[t]
  \centering
    \subfigure[]{\includegraphics[width=0.32\linewidth]{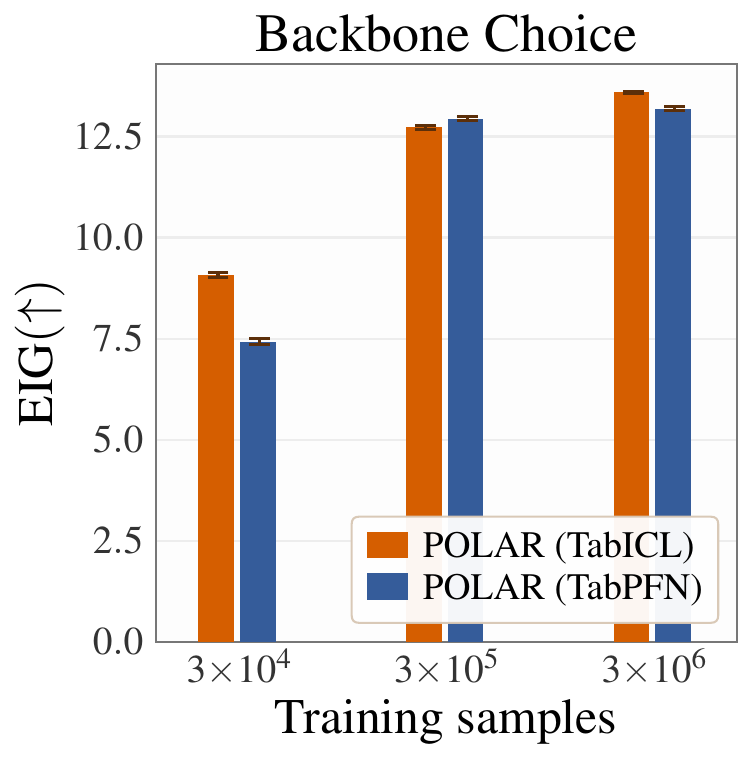}}
    \subfigure[]{\includegraphics[width=0.32\linewidth]{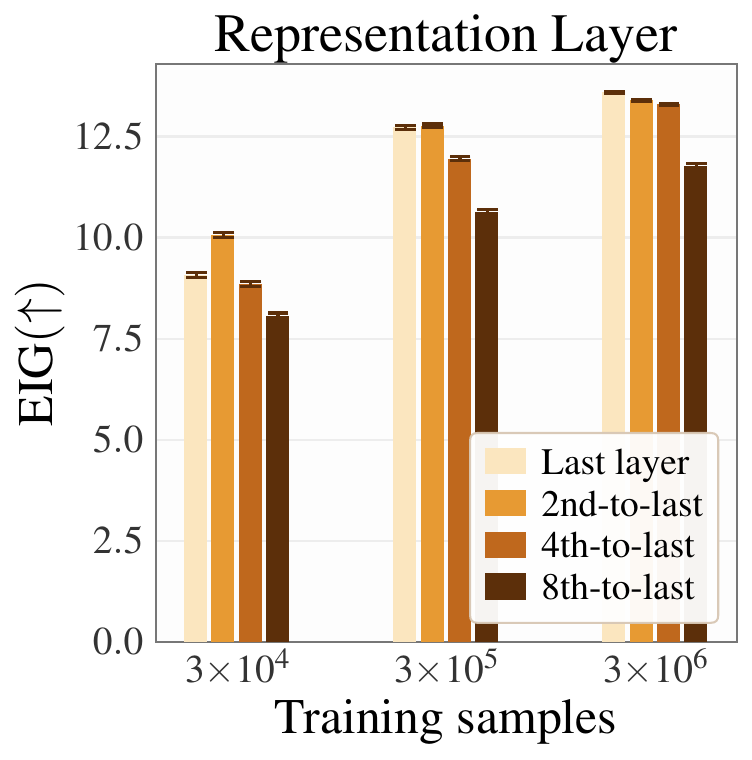}}
    \subfigure[]{\includegraphics[width=0.32\linewidth]{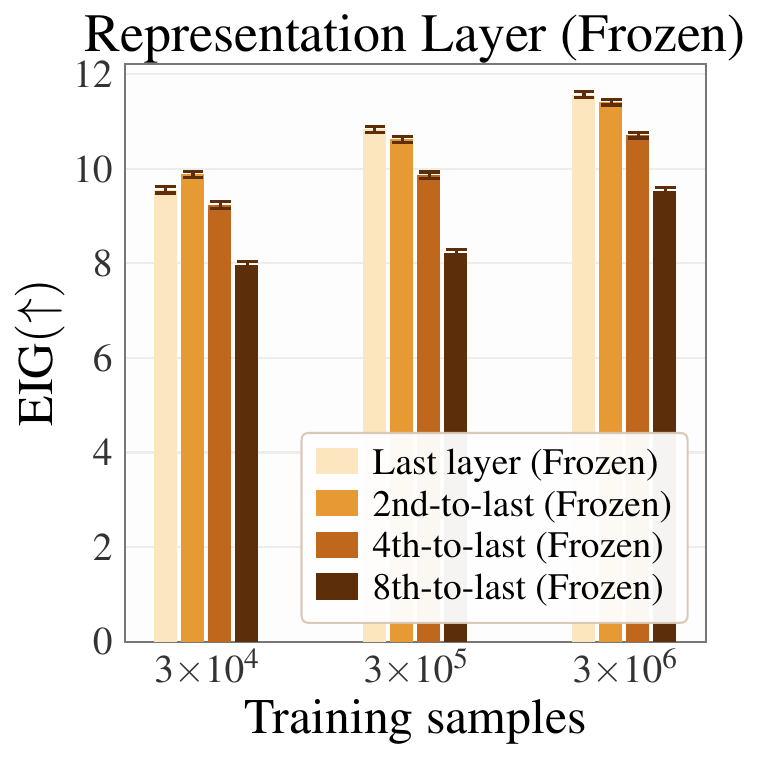}}
  \caption{\textbf{Location finding}. (a) Ablation study on the backbone choice. (b) Ablation study on the representation layer with backbone finetuning. (c) Ablation study on the representation layer with a frozen backbone.}
  \label{fig:location_finding_additional_ablations}
\end{figure}

\textbf{Backbone choice.} \Cref{fig:location_finding_additional_ablations}(a) examines whether \method's gains depend critically on the choice of pretrained backbone by replacing TabICLv2 with TabPFNv2.5 on the 2D task. The TabPFN-backed policy converges more slowly during the early stages of training, but the two variants reach essentially the same EIG at convergence. This suggests that \method is not tied to a particular backbone: as long as the pretrained model produces representations from which the belief can be approximately decoded, a lightweight policy head can turn them into an effective acquisition policy.

\textbf{Representation layer.} Another natural question is which layer of the backbone supplies the most informative belief representation for policy learning. To test this, we re-train \method on the 2D task using representations taken from the last, second-to-last, fourth-to-last, and eighth-to-last transformer layers of the TabICLv2 backbone, keeping all other components fixed. \Cref{fig:location_finding_additional_ablations}(b) shows the last and second-to-last layers perform comparably throughout training, with the second-to-last layer slightly ahead in the low-sample regime and the last layer marginally ahead at convergence. Earlier layers degrade performance monotonically. A natural concern is that the layer ordering in \Cref{fig:location_finding_additional_ablations}(b) reflects the effect of finetuning rather than the structure of the pretrained backbone. \Cref{fig:location_finding_additional_ablations}(c) repeats the comparison with a frozen backbone and shows the same ordering, confirming that later layers provide more informative belief representations regardless of whether the backbone is adapted. Therefore, we adopt the last layer as the default in all other experiments.

\begin{figure}[ht]
  \centering
    \includegraphics[width=0.7\linewidth]{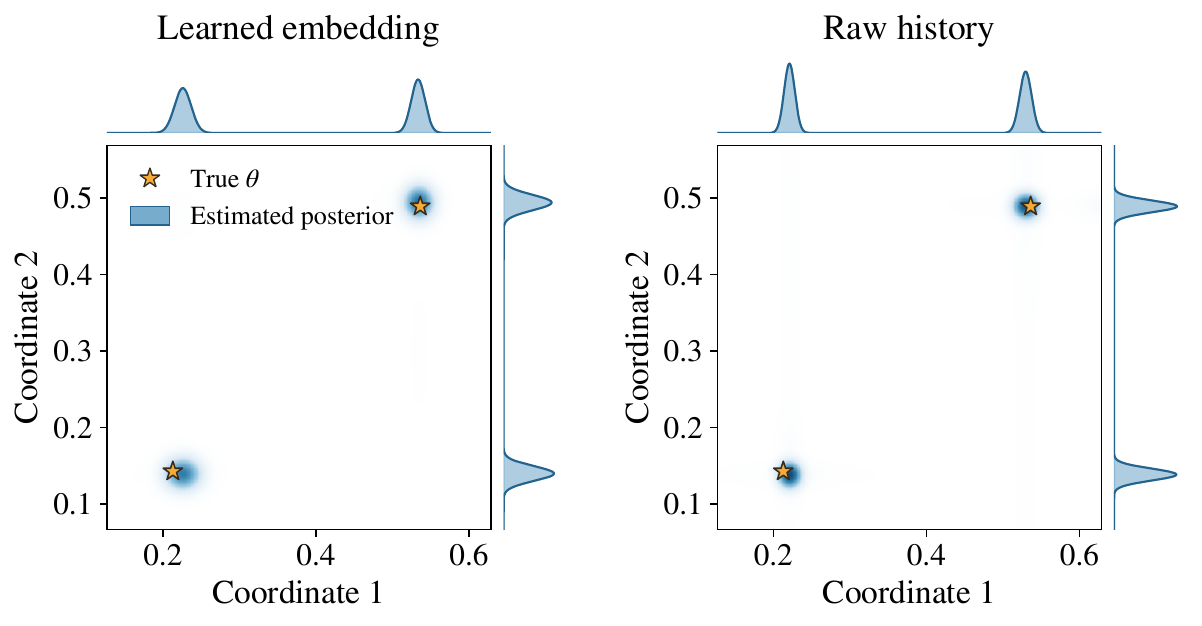
    }
  \caption{\textbf{Posterior decoding.} Factorised source-location densities from frozen \method embeddings (left) and raw history (right), for the same held-out trajectory after 30 acquisitions. Stars mark true source locations.}
  \label{fig:posterior_comparison}
\end{figure}

\textbf{Posterior decoding.} We investigate whether \method{}’s representations retain information about the latent parameters. After freezing the trained backbone and acquisition policy, we extract the final-layer history embeddings following 30 acquisitions. Four learned coordinate tokens cross-attend to these detached embeddings, and separate heads predict a 10-component Gaussian mixture for each coordinate, trained by marginal maximum likelihood. Sources are ordered by their first coordinate to resolve label-exchange ambiguity. As a control, a capacity-matched estimator encodes the corresponding raw histories with a small Transformer followed by an analogous marginal decoder, using the same trajectory split. On 2,000 held-out trajectories, both estimators achieve a posterior-mean RMSE of 0.04, compared with an expected RMSE of 0.26 for the canonicalised prior mean. \Cref{fig:posterior_comparison} shows a selected held-out example where both estimators localise the sources accurately. Together, these results provide empirical evidence that the frozen representations retain posterior information.

\textbf{Sensitivity to the training-score estimator.}
We vary the number of contrastive samples used in the sPCE training objective, $L_{\mathrm{train}}\in\{10^2,10^3,10^4,10^5\}$. \Cref{tab:contrastive_samples} shows that larger contrastive budgets improve policy learning at matched target-task sample budgets, with $10^4$ and $10^5$ attaining similar final performance.

\begin{table}[t]
    \centering
    \small
    \caption{Held-out sPCE (mean $\pm$ s.e.) on 2D location finding with different training contrastive-sample budgets.}
    \label{tab:contrastive_samples}
    \begin{tabular}{lcccc}
        \toprule
        Training samples
        & $L=10^2$
        & $L=10^3$
        & $L=10^4$
        & $L=10^5$ \\
        \midrule
        $3\times10^4$
        & $8.50\pm0.07$ & $8.55\pm0.07$
        & $9.46\pm0.07$ & $\mathbf{9.53}\pm0.06$ \\
        $3\times10^5$
        & $9.14\pm0.05$ & $9.80\pm0.06$
        & $12.09\pm0.06$ & $\mathbf{12.57}\pm0.05$ \\
        $3\times10^6$
        & $9.75\pm0.06$ & $12.67\pm0.06$
        & $13.58\pm0.02$ & $\mathbf{13.60}\pm0.02$ \\
        \bottomrule
    \end{tabular}
\end{table}

\subsection{Constant elasticity of substitution} \label{app:ces_exp}

\Cref{tab:ces_best} reports each method's best-attained performance. \method achieves the second strongest sPCE among all methods, close to the strongest baseline RL-BOED. During training, we observed that both DAD and RL-BOED were prone to instability: their training losses diverged in the late stages, with sPCE peaking and then degrading. We therefore applied early stopping based on validation sPCE for both baselines and report the checkpoint at which validation sPCE was highest.
A separate observation concerns the sPCE–sNMC gap. For RL-BOED, this gap is markedly larger than for the other methods. A wide gap at finite contrastive samples indicates that the EIG bounds are loose for this particular policy, rather than that the true EIG is correspondingly large. 

\begin{table}[h]
\centering
\caption{\textbf{CES}. Best performance comparison on the CES task. Results are reported as mean $\pm$ s.e. over 1,000 independent runs.}
\label{tab:ces_best}
\begin{tabular}{lcc}
\toprule
Methods &  sPCE & sNMC \\
\midrule
Random & \vpmse{9.18}{0.18} & \vpmse{11.86}{0.34} \\
DAD  & \vpmse{12.87}{0.14} & \vpmse{16.44}{0.39}  \\
RL-BOED & \vpmse{14.19}{0.09} & \vpmse{33.33}{1.80}  \\
ALINE &  \vpmse{13.43}{0.10} & \vpmse{18.13}{0.28} \\
\method  & \vpmse{14.11}{0.09} & \vpmse{20.09}{0.47}  \\
\bottomrule
\end{tabular}
\end{table}

\subsection{Hyperparameter optimisation} \label{app:hpo_exp}

\begin{figure}
  \centering
    \includegraphics[width=\linewidth]{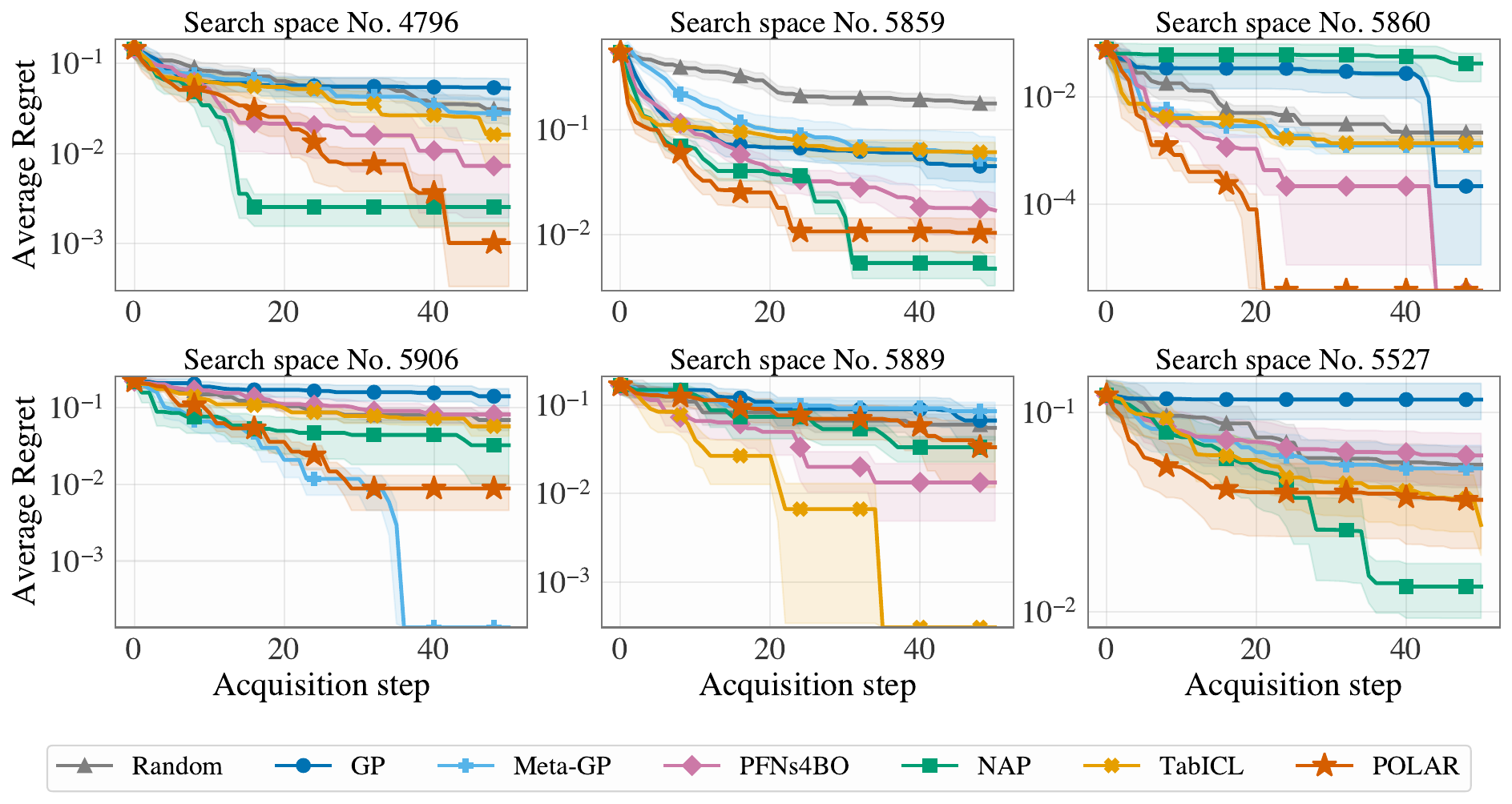}
  \caption{\textbf{HPO-B}. Per-search-space normalised regret over the acquisition trajectory. Shaded regions denote one standard error across test tasks and initialisation seeds.}
  \label{fig:hpo_per_search_space_regret}
\end{figure}

\begin{figure}
  \centering
    \includegraphics[width=\linewidth]{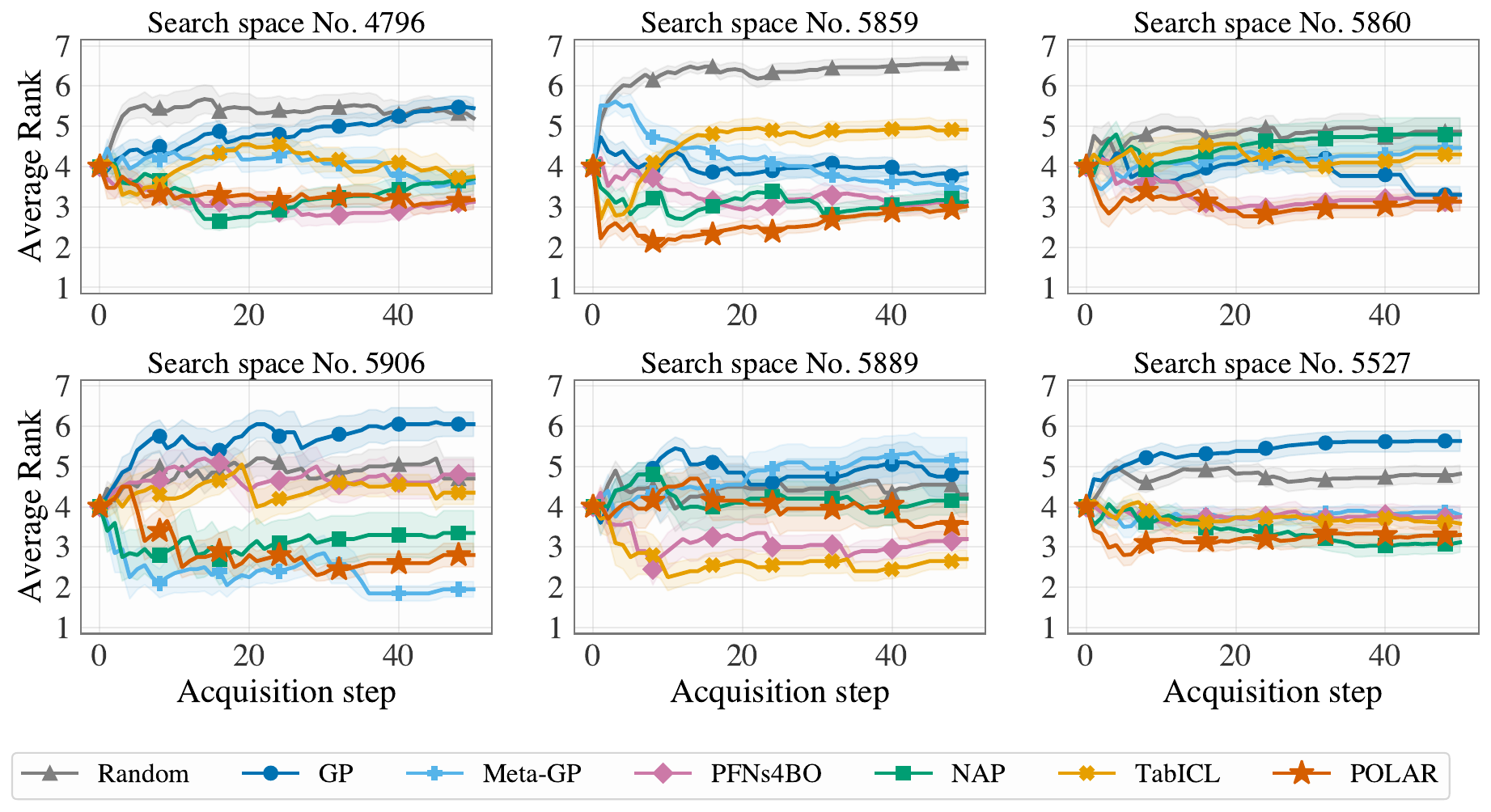}
  \caption{\textbf{HPO-B}. Per-search-space average rank over the acquisition trajectory. Shaded regions denote one standard error across test tasks and initialisation seeds.}
  \label{fig:hpo_per_search_space_rank}
\end{figure}

\Cref{fig:hpo_per_search_space_regret} and \Cref{fig:hpo_per_search_space_rank} break down the aggregate results of \Cref{sec:hpo} by search space, reporting the normalised regret and the average rank, respectively, for each of the six HPO-B search spaces. \method attains the lowest regret averaged over the acquisition trajectory on four of the six search spaces (4796, 5859, 5860, and 5527) and finishes in the top two on five of them, placing third on 5889. No single baseline matches this consistency: the best final-step method varies by search space, with NAP leading on 5859 and 5527, Meta-GP on 5906, and TabICL on 5889.

\subsection{Role of pretrained belief representations} \label{app:random_backbone}

In \Cref{sec:bed_exp}, we compare pretrained frozen, pretrained adapted (\method), and randomly initialised trainable backbones. These experiments establish the benefits of pretraining and task-specific adaptation, but do not directly test whether the policy head can learn an effective acquisition rule from fixed random features. To further isolate the contribution of the pretrained belief representation, we replace the policy's backbone with an architecturally identical, randomly initialised frozen backbone while keeping the policy head and training procedure unchanged. We perform this control in two settings: loss-driven active learning (\Cref{app:al_exp}) with the target-aware transformer head, and synthetic meta-BO with the lightweight MLP head. 

\textbf{Loss-driven active learning.} We follow the setup in \Cref{app:al_exp}, using the target-aware Transformer policy head. Since the backbone also supplies the predictive distributions used to estimate rewards, we use a dual-backbone control: a pretrained backbone continues to provide predictive rewards and evaluation, while the policy head reads candidate and target representations from the frozen random encoder. This separates the quality of the policy's input representation from the validity of reward estimation. \Cref{tab:random_backbone_al} shows substantial degradation on both in-distribution GP functions and held-out benchmark functions.

\begin{table}[t]
    \centering
    \small
    \caption{Random-backbone control for loss-driven AL.
    Final-step weighted prediction errors (mean $\pm$ s.e.; lower is better).}
    \label{tab:random_backbone_al}
    \begin{tabular}{lcccc}
        \toprule
        & \multicolumn{2}{c}{GP synthetic}
        & \multicolumn{2}{c}{Benchmark functions} \\
        \cmidrule(lr){2-3}\cmidrule(lr){4-5}
        Method
        & $\mathrm{RMSE}_{\omega} (\downarrow)$ & $\mathrm{NLPD}_{\omega} (\downarrow)$
        & $\mathrm{RMSE}_{\omega} (\downarrow)$ & $\mathrm{NLPD}_{\omega} (\downarrow)$ \\
        \midrule
        \method
        & $\mathbf{0.09}\pm0.03$ & $\mathbf{-2.07}\pm0.13$
        & $\mathbf{0.05}\pm0.01$ & $\mathbf{-1.68}\pm0.10$ \\
        Random backbone
        & $0.56\pm0.07$ & $0.16\pm0.21$
        & $0.30\pm0.05$ & $0.27\pm0.22$ \\
        \bottomrule
    \end{tabular}
\end{table}

\textbf{Synthetic meta-BO.}
We additionally test the standard two-hidden-layer MLP policy head on a meta-BO task. Here, meta-BO \citep{maraval2023end} means learning a single acquisition policy across a distribution of optimisation tasks and deploying it on unseen functions without further policy training. Each training episode samples a new function using a two-dimensional variant of the synthetic-GP sampler used for loss-driven AL in \Cref{app:al_exp}. Unlike loss-driven AL, here the reward is the improvement in the best observed value, equivalently the reduction in simple regret, normalised by the episode's objective range.

Both policies use the same two-hidden-layer MLP head and are trained for 5,000 updates with batches of 32 trajectories and 30 acquisitions per trajectory. We evaluate on 50 new GP functions and four standard 2D benchmarks: Branin, Himmelblau, Styblinski--Tang, and McCormick, with 20 initialisations per benchmark. Each evaluation trajectory starts from five observations and uses a candidate pool of 2,048 points. \Cref{tab:random_backbone_bo} reports final simple regret on standardised objectives after 30 acquisitions.

\begin{table}[t]
    \centering
    \small
    \caption{Random-backbone control for the MLP BO policy.
    Final simple regret on standardised objectives
    (mean $\pm$ s.e.; lower is better).}
    \label{tab:random_backbone_bo}
    \begin{tabular}{lcc}
        \toprule
        Method & GP synthetic & Benchmark functions \\
        \midrule
        \method
        & $\mathbf{0.075}\pm0.024$
        & $\mathbf{0.011}\pm0.002$ \\
        Random backbone
        & $0.545\pm0.077$
        & $0.248\pm0.032$ \\
        \bottomrule
    \end{tabular}
\end{table}

In both settings, the random-backbone policy performs substantially worse than \method, showing that neither head can learn an effective acquisition rule on its own and that the pretrained belief representation contributes materially to performance.

\subsection{Controlled distribution shift}\label{app:controlled_shift}

To test the robustness of \method, we study a simple, controlled shift between policy training and deployment. We use ordinary, unweighted active learning on one-dimensional RBF GP functions, training separate policies under lengthscale priors $\ell\sim\mathcal{U}(0.1,0.8)$ and $\ell\sim\mathcal{U}(0.4,0.8)$. Each trajectory starts from two observations and selects 20 acquisitions from a pool of 200 candidates, with 100 target points. We retain the target-aware policy head and auxiliary prediction loss. Evaluation uses unweighted RMSE.

Each trained policy is evaluated at fixed lengthscales $\ell\in\{0.1,0.2,0.3,0.4,0.6,0.8\}$. As shown in \Cref{tab:controlled_shift}, the restricted-prior policy performs comparably to the broad-prior policy within its training support and incurs only a small additional error outside it. Importantly, shorter-lengthscale functions are harder for both policies: the modest mismatch penalty refers to the difference between policies at the same test lengthscale, not to a small absolute increase in RMSE. This experiment provides evidence of graceful transfer under the tested parametric shift within a fixed kernel family; we do not claim robustness to arbitrary out-of-distribution tasks.

\begin{table}[t]
    \centering
    \small
    \caption{Controlled lengthscale shift in ordinary AL.
    Final unweighted RMSE (mean $\pm$ s.e.; lower is better).}
    \label{tab:controlled_shift}
    \begin{tabular}{lcccccc}
        \toprule
        Training prior
        & $\ell=0.1$ & $\ell=0.2$ & $\ell=0.3$
        & $\ell=0.4$ & $\ell=0.6$ & $\ell=0.8$ \\
        \midrule
        $\mathcal{U}(0.1,0.8)$
        & $0.45\pm0.01$ & $0.09\pm0.01$ & $0.02\pm0.00$
        & $0.01\pm0.00$ & $0.01\pm0.00$ & $0.01\pm0.00$ \\
        $\mathcal{U}(0.4,0.8)$
        & $0.48\pm0.02$ & $0.10\pm0.01$ & $0.03\pm0.00$
        & $0.01\pm0.00$ & $0.01\pm0.00$ & $0.01\pm0.00$ \\
        \bottomrule
    \end{tabular}
\end{table}

\subsection{Stochastic pendulum}
\label{app:pendulum}

We study whether one-step utility provides sufficient training credit when actions influence the informativeness of future observations.

\textbf{Task setup.}
We adopt the stochastic single-pendulum setup used in Action-BED \citep{rossa2026action}, based on the conditionally linear model of \citet{iqbal2024nesting}. The observed state consists of angular position $q_t$ and velocity $v_t=\dot q_t$, and the design is an applied torque $x_t\in[-1,1]$. The discretised dynamics are
\begin{equation}
\begin{aligned}
    q_{t+1}
    &=q_t+\Delta t\,v_t,\\
    v_{t+1}
    &=v_t+\Delta t
    \left[-\theta_1\sin(q_t)-\theta_2v_t+\theta_3 x_t\right]
    +0.1\sqrt{\Delta t}\,\epsilon_t,
\end{aligned}
\label{eq:pendulum_dynamics}
\end{equation}
where $\epsilon_t\sim\mathcal N(0,1)$ independently, $\Delta t=0.05$, and the horizon is $T=50$. Each trajectory starts at $(q_0,v_0)=(0,0)$. The unknown restoring, damping, and torque-response coefficients have prior
\[
    \theta\sim\mathcal N\!\left(
        (14.7,0,3)^\top,\,
        \operatorname{diag}(0.1,0.01,0.1)
    \right).
\]

At the initial state, the immediate velocity increment is insensitive to $\theta_1$ and $\theta_2$. Earlier torques must first excite the position and velocity before subsequent observations become informative about these coefficients. An action can therefore have limited immediate utility while improving future measurement opportunities.

\textbf{Implementation.}
The backbone receives history rows $x_t=(q_t,v_t,x_t)$ and observations $y_t=v_{t+1}-v_t$. At each step, a two-hidden-layer MLP with width 128 and GELU activations scores candidate-conditioned TFM representations, augmented with the candidate torque and remaining-horizon fraction. Training uses 101 uniformly spaced torque candidates. 

\textbf{Training and evaluation.}
We compare one-step credit ($\gamma=0$) with undiscounted reward-to-go ($\gamma=1$) in \Cref{eq:policy_loss}, keeping the simulator, architecture, training budget, and other hyperparameters fixed. Both variants use the same dense sPCE utility increments, computed with $L=10{,}000$ contrastive samples. Each policy is trained for 5,000 updates with batches of 64 trajectories. The policy-head and backbone learning rates are $3\times10^{-4}$ and $10^{-5}$, respectively. 

For final evaluation, each policy uses its own adapted backbone and selects torques greedily from 201 uniformly spaced candidates. We report the terminal sPCE bound on 2,000 independent trajectories with $L=50{,}000$ contrastive samples. The random baseline samples torques uniformly from $[-1,1]$.

\textbf{Results.}
\Cref{tab:pendulum_credit} shows that reward-to-go substantially outperforms one-step credit.
The utility definition and model architecture are unchanged: the difference is whether later informational benefits are credited to earlier actions. This supports choosing the credit-assignment horizon according to the task's temporal structure. One-step credit is effective in our original benchmarks, whereas reward-to-go is preferable in this dynamical task with delayed informational consequences.

\begin{table}[t]
    \centering
    \small
    \caption{Temporal credit assignment on stochastic-pendulum BED.
    Terminal sPCE (mean $\pm$ s.e.; higher is better).}
    \label{tab:pendulum_credit}
    \begin{tabular}{lc}
        \toprule
        Method & sPCE \\
        \midrule
        Random & $1.39\pm0.03$ \\
        \method, one-step credit ($\gamma=0$)
        & $2.39\pm0.03$ \\
        \method, reward-to-go ($\gamma=1$)
        & $\mathbf{3.96}\pm0.04$ \\
        \bottomrule
    \end{tabular}
\end{table}

\section{Computational resources and software} \label{app:resource_and_license}

All experiments presented in this work, including model development, hyperparameter tuning, baseline evaluations, and preliminary analyses, were performed on GPU clusters equipped with NVIDIA H200 and AMD MI250x GPUs. The total computational resources consumed for this research, across all development stages and experimental runs, are estimated to be approximately 2000 GPU hours. The wall-clock cost of training \method varies with the task: on the HPO-B benchmark, a strong policy is typically obtained within 30 minutes per search space; on the location finding, policy training takes around 5 hours; on the CES, it takes about 16 hours to get the final best performance; on the \textsc{Dockstring} benchmark, the high input dimensionality increases the per-step cost of the backbone forward pass, and a full training run takes approximately one day.
The core code base is implemented in PyTorch (\url{https://pytorch.org}, License: modified BSD license), which also underlies our policy head as well as the TabPFN and TabICL backbones. We use the publicly released TabPFNv2.5 backbone from the PriorLabs repository (\url{https://github.com/PriorLabs/TabPFN}, License: Apache 2.0), with the corresponding checkpoint hosted on Hugging Face (\url{https://huggingface.co/Prior-Labs/tabpfn_2_5}). For the TabICLv2 backbone, we use the official implementation released by the authors (\url{https://github.com/soda-inria/tabicl}, License: BSD 3-Clause), with the regression checkpoint hosted on Hugging Face (\url{https://huggingface.co/jingang/TabICL}).
BED baselines are adapted from the original authors' publicly available code. For the HPO-B baselines, we use the publicly released hebo-plus-model checkpoint from the official PFNs4BO repository (\url{https://github.com/automl/PFNs4BO}, License: Apache 2.0), and we use the per-step regret values released as part of the official NAP codebase (\url{https://github.com/huawei-noah/HEBO/tree/master/NAP}) for the NAP comparison. The Gaussian process baselines are implemented using GPyTorch (\url{https://github.com/cornellius-gp/gpytorch}, License: MIT). The HPO-B benchmark is obtained from its official repository (\url{https://github.com/releaunifreiburg/HPO-B}, License: MIT), and the \textsc{Dockstring} benchmark is obtained from its official repository (\url{https://github.com/dockstring/dockstring}, License: Apache 2.0).


\end{document}